\documentclass[10pt,journal,compsoc]{IEEEtran}

\ifCLASSOPTIONcompsoc
  \usepackage[nocompress]{cite}
\else
  \usepackage{cite}
\fi

\ifCLASSINFOpdf
\else
\fi

\usepackage{amsmath}
\usepackage[none]{hyphenat}
\usepackage{epsfig,graphicx,subfigure}
\usepackage{multirow}
\usepackage{multicol}
\usepackage{amssymb}
\usepackage{amsmath}
\usepackage{amsfonts}
\usepackage{subfigure}
\usepackage{graphicx}
\usepackage{multirow}
\usepackage{url}
\usepackage{paralist}
\usepackage{mathrsfs, euscript}
\usepackage{color}
\usepackage{pifont}

\begin{document}

\def\Blue{\color{blue}}
\def\Purple{\color{purple}}

\def\A{{\bf A}}
\def\a{{\bf a}}
\def\B{{\bf B}}
\def\b{{\bf b}}
\def\C{{\bf C}}
\def\c{{\bf c}}
\def\D{{\bf D}}
\def\d{{\bf d}}
\def\E{{\bf E}}
\def\e{{\bf e}}
\def\f{{\bf f}}
\def\F{{\bf F}}
\def\K{{\bf K}}
\def\k{{\bf k}}
\def\L{{\bf L}}
\def\H{{\bf H}}
\def\h{{\bf h}}
\def\G{{\bf G}}
\def\g{{\bf g}}
\def\I{{\bf I}}
\def\J{{\bf J}}
\def\R{{\bf R}}
\def\X{{\bf X}}
\def\Y{{\bf Y}}
\def\OO{{\bf O}}
\def\oo{{\bf o}}
\def\P{{\bf P}}
\def\p{{\bf p}}
\def\Q{{\bf Q}}
\def\q{{\bf q}}
\def\r{{\bf r}}
\def\s{{\bf s}}
\def\S{{\bf S}}
\def\t{{\bf t}}
\def\T{{\bf T}}
\def\x{{\bf x}}
\def\y{{\bf y}}
\def\z{{\bf z}}
\def\Z{{\bf Z}}
\def\M{{\bf M}}
\def\m{{\bf m}}
\def\n{{\bf n}}
\def\U{{\bf U}}
\def\u{{\bf u}}
\def\V{{\bf V}}
\def\v{{\bf v}}
\def\W{{\bf W}}
\def\w{{\bf w}}
\def\0{{\bf 0}}
\def\1{{\bf 1}}

\def\AM{{\mathcal A}}
\def\EM{{\mathcal E}}
\def\FM{{\mathcal F}}
\def\TM{{\mathcal T}}
\def\UM{{\mathcal U}}
\def\XM{{\mathcal X}}
\def\YM{{\mathcal Y}}
\def\NM{{\mathcal N}}
\def\OM{{\mathcal O}}
\def\IM{{\mathcal I}}
\def\GM{{\mathcal G}}
\def\PM{{\mathcal P}}
\def\LM{{\mathcal L}}
\def\MM{{\mathcal M}}
\def\DM{{\mathcal D}}
\def\SM{{\mathcal S}}
\def\RB{{\mathbb R}}
\def\EB{{\mathbb E}}

\def\tx{\tilde{\bf x}}
\def\ty{\tilde{\bf y}}
\def\tz{\tilde{\bf z}}
\def\hd{\hat{d}}
\def\HD{\hat{\bf D}}
\def\hx{\hat{\bf x}}
\def\hR{\hat{R}}

\def\Ome{\mbox{\boldmath$\omega$\unboldmath}}
\def\Om{\mbox{\boldmath$\Omega$\unboldmath}}
\def\bet{\mbox{\boldmath$\beta$\unboldmath}}
\def\et{\mbox{\boldmath$\eta$\unboldmath}}
\def\ep{\mbox{\boldmath$\epsilon$\unboldmath}}
\def\ph{\mbox{\boldmath$\phi$\unboldmath}}
\def\Pii{\mbox{\boldmath$\Pi$\unboldmath}}
\def\pii{\mbox{\boldmath$\pi$\unboldmath}}
\def\Ph{\mbox{\boldmath$\Phi$\unboldmath}}
\def\Ps{\mbox{\boldmath$\Psi$\unboldmath}}
\def\tha{\mbox{\boldmath$\theta$\unboldmath}}
\def\Tha{\mbox{\boldmath$\Theta$\unboldmath}}
\def\muu{\mbox{\boldmath$\mu$\unboldmath}}
\def\Si{\mbox{\boldmath$\Sigma$\unboldmath}}
\def\si{\mbox{\boldmath$\sigma$\unboldmath}}
\def\Gam{\mbox{\boldmath$\Gamma$\unboldmath}}
\def\gamm{\mbox{\boldmath$\gamma$\unboldmath}}
\def\Lam{\mbox{\boldmath$\Lambda$\unboldmath}}
\def\De{\mbox{\boldmath$\Delta$\unboldmath}}
\def\vps{\mbox{\boldmath$\varepsilon$\unboldmath}}
\def\Up{\mbox{\boldmath$\Upsilon$\unboldmath}}
\def\xii{\mbox{\boldmath$\xi$\unboldmath}}
\def\Xii{\mbox{\boldmath$\Xi$\unboldmath}}
\def\Lap{\mbox{\boldmath$\LM$\unboldmath}}
\newcommand{\ti}[1]{\tilde{#1}}

\def\tr{\mathrm{tr}}
\def\etr{\mathrm{etr}}
\def\etal{{\em et al.\/}\,}
\newcommand{\indep}{{\;\bot\!\!\!\!\!\!\bot\;}}
\def\argmax{\mathop{\rm argmax}}
\def\argmin{\mathop{\rm argmin}}
\def\vec{\text{vec}}
\def\cov{\text{cov}}
\def\dg{\text{diag}}

\title{Towards Bayesian Deep Learning:\\A Framework and Some Existing Methods}

\author{Hao Wang,
        Dit-Yan Yeung~\IEEEmembership{Senior Member,~IEEE}\\% <-this % stops a space
        %Hong Kong University of Science and Technology\\
        %\tt \{hwangaz,dyyeung\}@cse.ust.hk
\IEEEcompsocitemizethanks{\IEEEcompsocthanksitem Hao Wang is with the Department of Computer Science and Engineering, the Hong Kong University of Science and Technology. \protect
%% note need leading \protect in front of \\ to get a newline within \thanks as
%% \\ is fragile and will error, could use \hfil\break instead.
E-mail: hwangaz@cse.ust.hk
\IEEEcompsocthanksitem Dit-Yan Yeung is with the Department of Computer Science and Engineering, the Hong Kong University of Science and Technology. \protect
E-mail: dyyeung@cse.ust.hk}% <-this % stops an unwanted space
}

\markboth{IEEE TRANSACTIONS ON KNOWLEDGE AND DATA ENGINEERING}%
{Shell \MakeLowercase{\textit{et al.}}: Bare Demo of IEEEtran.cls for Computer Society Journals}




\IEEEtitleabstractindextext{%
\begin{abstract}
While perception tasks such as visual object recognition and text understanding play an important role in human intelligence, subsequent tasks that involve inference, reasoning and planning require an even higher level of intelligence.  The past few years have seen major advances in many perception tasks using deep learning models.  For higher-level inference, however, probabilistic graphical models with their Bayesian nature are still more powerful and flexible.  To achieve integrated intelligence that involves both perception and inference, it is naturally desirable to tightly integrate deep learning and Bayesian models within a principled probabilistic framework, which we call \emph{Bayesian deep learning}. In this unified framework, the perception of text or images using deep learning can boost the performance of higher-level inference and in return, the feedback from the inference process is able to enhance the perception of text or images. This paper proposes a general framework for \emph{Bayesian deep learning} and reviews its recent applications on recommender systems, topic models, and control. In this paper, we also discuss the relationship and differences between Bayesian deep learning and other related topics such as the Bayesian treatment of neural networks.
\end{abstract}

}

\maketitle

\IEEEdisplaynontitleabstractindextext

% \fi
\IEEEpeerreviewmaketitle

\IEEEraisesectionheading{\section{Introduction}\label{sec:intro}}

\IEEEPARstart{D}{eep} learning has achieved significant success in many perception tasks including \emph{seeing} (visual object recognition), \emph{reading} (text understanding), and \emph{hearing} (speech recognition). These are undoubtedly fundamental tasks for a functioning comprehensive artificial intelligence (AI) or data engineering (DE) system. However, in order to build a real AI/DE system, simply being able to see, read, and hear is far from enough. It should, above all, possess the ability to \emph{think}.

Take medical diagnosis as an example. Besides \emph{seeing} visible symptoms (or medical images from CT) and \emph{hearing} descriptions from patients, a doctor has to look for relations among all the symptoms and preferably infer the corresponding etiology. Only after that can the doctor provide medical advice for the patients. In this example, although the abilities of \emph{seeing} and \emph{hearing} allow the doctor to acquire information from the patients, it is the \emph{thinking} part that defines a doctor. Specifically, the ability to \emph{think} here could involve causal inference, logic deduction, and dealing with uncertainty, which is apparently beyond the capability of conventional deep learning methods. Fortunately, another type of models, probabilistic graphical models (PGM), excels at causal inference and dealing with uncertainty. The problem is that PGM is not as good as deep learning models at perception tasks. To address the problem, it is, therefore, a natural choice to tightly integrate deep learning and PGM within a principled probabilistic framework, which we call \emph{Bayesian deep learning} (BDL) in this paper.

With the tight and principled integration in BDL, perception tasks and inference tasks are regarded as a whole and can benefit from each other. In the example above, being able to see the medical image could help with the doctor's diagnosis and inference. On the other hand, diagnosis and inference can in return help with understanding the medical image. Suppose a doctor is not sure what a dark spot in a medical image is. However, if she is able to \emph{infer} the etiology of the symptoms and disease, it can help her better decide whether the dark spot is a tumor or not.

As another example, to achieve high accuracy in recommender systems (RS) \cite{CDL,DBLP:conf/aaai/LuDLX015,DBLP:journals/tkde/AdomaviciusK12,ricci2011introduction,DBLP:conf/recsys/LiuMLY11}, we need to fully understand the content of the items (e.g., documents and movies) \cite{DBLP:journals/tkde/Park13}, analyze the profile and preferences of users \cite{DBLP:journals/tkde/WeiMJ05,DBLP:conf/aaai/ZhengCZXY10}, and evaluate the similarity among the users \cite{DBLP:journals/tkde/CaiLLMTL14,DBLP:journals/tkde/HornickT12,DBLP:journals/tkde/BartoliniZP11}. Deep learning is good at the first subtask while PGM excels at the other two. Besides the fact that better understanding of item content would help with the analysis of user profiles, the estimated similarity among users could also provide valuable information for understanding item content in return. In order to fully utilize this bidirectional effect to boost recommendation accuracy, we might wish to unify deep learning and PGM in one single principled probabilistic framework, as seen in \cite{CDL}.

Besides recommender systems, the need for BDL may also arise when we are dealing with the control of non-linear dynamic systems with raw images as input. Consider controlling a complex dynamical system according to the live video stream received from a camera. This problem can be transformed into iteratively performing two tasks, \emph{perception} from raw images and \emph{control} based on dynamic models. The perception task can be taken care of using multiple layers of simple nonlinear transformation (deep learning) while the control task usually needs more sophisticated models like hidden Markov models and Kalman filters \cite{harrison1999bayesian}. The feedback loop is then completed by the fact that actions chosen by the control model can affect the received video stream in return. To enable an effective iterative process between the perception task and the control task, we need two-way information exchange between them. The perception component would be the basis on which the control component estimates its states and the control component with a built-in dynamic model would be able to predict the future trajectory (images). In such cases, BDL is a suitable choice \cite{watter2015embed}.

As mentioned in the examples above, BDL is particularly useful for tasks that involve both understanding of content (e.g., text, images, and videos) and inference/reasoning among variables. In such complex tasks, the perception component of BDL is responsible for the understanding of the content, and the task-specific component (e.g., the control component in dynamical systems) models the probabilistic relationship among different variables. Furthermore, the interaction between these two components creates synergy and further boosts the performance.

Apart from the major advantage of BDL providing a principled way of unifying deep learning and PGM, another benefit comes from the implicit regularization built into BDL. Through imposing a prior on hidden units, parameters defining a neural network, or the model parameters specifying the causal inference, to some degree BDL can avoid overfitting, especially when there is not sufficient data. Usually, a BDL model consists of two components: (1) a \emph{perception component} that is a Bayesian formulation of a certain type of neural networks and (2) a \emph{task-specific component} that describes the relationship among different hidden or observed variables using PGM. Regularization is crucial for them both. Neural networks usually have large numbers of free parameters that need to be regularized properly. Regularization techniques such as weight decay and dropout \cite{srivastava2014dropout} are shown to be effective in improving performance of neural networks and they both have Bayesian interpretations \cite{gal2015dropout}. In terms of the task-specific component, expert knowledge or prior information, as a kind of regularization, can be incorporated into the model through the prior we imposed to guide the model when data are scarce.

Yet another advantage of using BDL for complex tasks (tasks that need both perception and inference) is that it provides a principled Bayesian approach of handling parameter uncertainty. When BDL is applied to complex tasks, there are \emph{three kinds of parameter uncertainties} that need to be taken into account:
\begin{compactenum}
 \item Uncertainty about the neural network parameters.
 \item Uncertainty about the task-specific parameters.
 \item Uncertainty about the exchange of information between the perception component and the task-specific component.
\end{compactenum}
Through representing the unknown parameters using distributions instead of point estimates, BDL offers a promising framework to handle these three kinds of uncertainty in a unified way. It is worth noting that the third uncertainty could only be handled under a unified framework such as BDL. If we train the perception component and the task-specific component separately, it is equivalent to assuming \emph{no uncertainty} when \emph{exchanging information} between the two components.

Of course, there are challenges when applying BDL to real-world tasks. (1) First, it is nontrivial to design an efficient Bayesian formulation of neural networks with reasonable time complexity. This line of work has been pioneered by \cite{mackay1992practical,hinton1993keeping,neal1995bayesian}, but it has not been widely adopted due to its lack of scalability. Fortunately, some recent advances in this direction \cite{kingma2013auto,DBLP:conf/icml/Hernandez-Lobato15b,DBLP:conf/icml/BlundellCKW15,balan2015bayesian,NPN} seem to shed light on the practical adoption of Bayesian neural networks\footnote{Here we refer to Bayesian treatment of neural networks as Bayesian neural networks. The other term, Bayesian deep learning, is retained to refer to complex Bayesian models with both a perception component and a task-specific component.}. (2) The second challenge is to ensure efficient and effective information exchange between the perception component and the task-specific component. Ideally both the first-order and second-order information (e.g., the mean and the variance) should be able to flow back and forth between the two components. A natural way is to represent the perception component as a PGM and seamlessly connect it to the task-specific PGM, as done in \cite{RSDAE,CDL,DPFA}.

In this paper, we aim to give a comprehensive overview of BDL models for applications like recommender systems, topic models (and representation learning), and control. The rest of the paper is organized as follows: In Section \ref{sec:dl}, we provide a review of some basic deep learning models. Section \ref{sec:pgm} covers the main concepts and techniques for PGM. These two sections serve as the background for BDL, and the next section, Section \ref{sec:bdl}, proposes a unified BDL framework and surveys the BDL models applied to areas such as recommender systems and topic models. Section \ref{sec:summary} discusses some future research issues and concludes the paper.

\section{Deep Learning}\label{sec:dl}
Deep learning normally refers to neural networks with more than two layers. To better understand deep learning, here we start with the simplest type of neural networks, multilayer perceptrons (MLP), as an example to show how conventional deep learning works. After that, we will review several other types of deep learning models based on MLP.

\subsection{Multilayer Perceptron}
Essentially a multilayer perceptron is a sequence of parametric nonlinear transformations. Suppose we want to train a multilayer perceptron to perform a regression task which maps a vector of $M$ dimensions to a vector of $D$ dimensions. We denote the input as a matrix $\X_0$ ($0$ means it is the $0$-th layer of the perceptron). The $j$-th row of $\X_0$, denoted as $\X_{0,j*}$, is an $M$-dimensional vector representing one data point. The target (the output we want to fit) is denoted as $\Y$. Similarly $\Y_{j*}$ denotes a $D$-dimensional row vector. The problem of learning an $L$-layer multilayer perceptron can be formulated as the following optimization problem:

\begin{align*}
\min\limits_{\{\W_l\},\{\b_l\}} ~&\|\X_L-\Y\|_F+\lambda\sum\limits_l\|\W_l\|_F^2\\
\mbox{subject to}~~&\X_{l}=\sigma(\X_{l-1}\W_l+\b_l), l=1,\dots,L-1\\
&\X_{L}=\X_{L-1}\W_L+\b_L,
\end{align*}
where $\sigma(\cdot)$ is an element-wise sigmoid function for a matrix and $\sigma(x)=\frac{1}{1+\exp(-x)}$. $\lambda$ is a regularization parameter and $\|\cdot\|_F$ denotes the Frobenius norm. The purpose of imposing $\sigma(\cdot)$ is to allow nonlinear transformation. Normally other transformations like $\tanh(x)$ and $\max(0,x)$ can be used as alternatives of the sigmoid function.

Here $\X_l$ ($l=1,2,\dots,L-1$) is the hidden units. As we can see, $\X_L$ can be easily computed once $\X_0$, $\W_l$, and $\b_l$ are given. Since $\X_0$ is given by the data, we only need to learn $\W_l$ and $\b_l$ here. Usually this is done using backpropagation and stochastic gradient descent (SGD). The key is to compute the gradients of the objective function with respect to $\W_l$ and $\b_l$. If we denote the value of the objective function as $E$, we can compute the gradients using the chain rule as:
\begin{align}
\frac{\partial E}{\partial \X_L}&=2(\X_L-\Y)\label{eq:grad_con_L}\\
\frac{\partial E}{\partial \X_l}&=(\frac{\partial E}{\partial \X_{l+1}}\circ\X_{l+1}\circ(1-\X_{l+1}))\W_{l+1}\label{eq:grad_con_all}\\
\frac{\partial E}{\partial \W_l}&=\X_{l-1}^T(\frac{\partial E}{\partial \X_l}\circ\X_l\circ(1-\X_l))\label{eq:grad_W_con}\\
\frac{\partial E}{\partial \b_l}&=mean(\frac{\partial E}{\partial \X_l}\circ\X_l\circ(1-\X_l),1),\label{eq:grad_b_con}
\end{align}
where $l=1,\dots,L$ and the regularization terms are omitted. The element-wise product is denoted as $\circ$ and $mean(\cdot,1)$ is the matlab operation on matrices. In practice, we only use a small part of the data (e.g., $128$ data points) to compute the gradients for each update. This is called stochastic gradient descent.

As we can see, in conventional deep learning models, only $\W_l$ and $\b_l$ are free parameters, which we will update in each iteration of the optimization. $\X_l$ is not a free parameter since it can be computed exactly if $\W_l$ and $\b_l$ are given.

\begin{figure}[!tb]
\begin{center}
%\framebox[4.0in]{$\;$}
%\includegraphics[height=5cm]{likeli1.eps}
\subfigure{
\includegraphics[height=3.0cm]{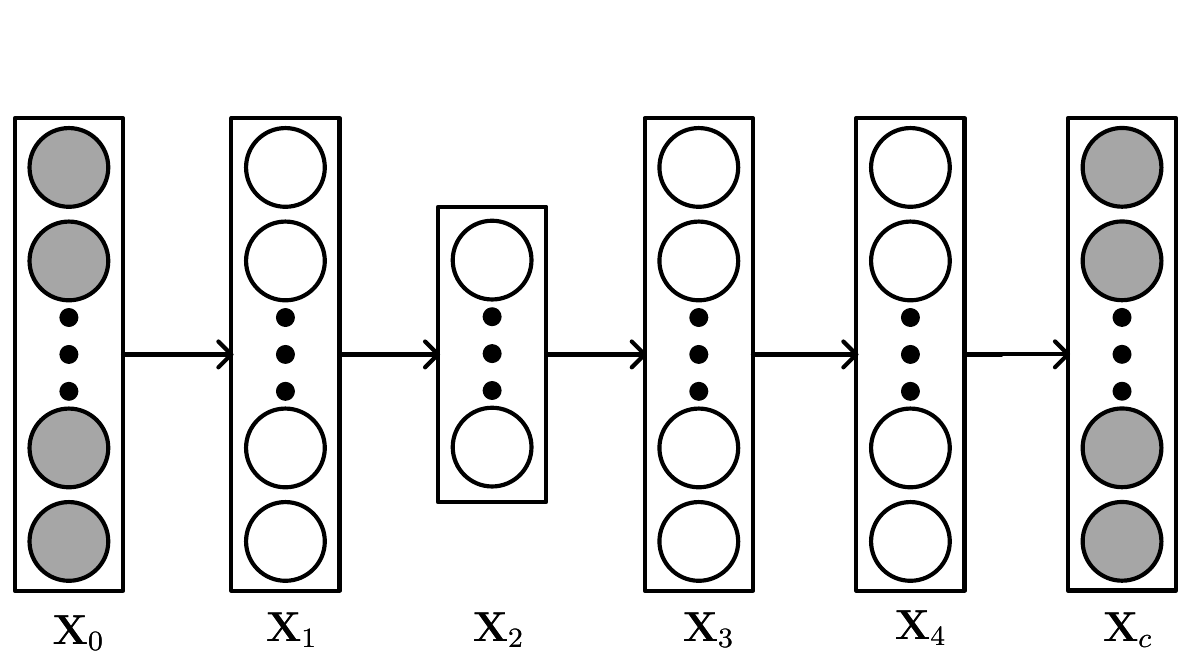}}
\end{center}
\vskip -0.2in
\caption{A 2-layer SDAE with $L=4$.
%$P=1$ is the sparse setting and $P=10$ is the dense setting.  $P$ is the number of ratings for each user in the training set.
}
\label{fig:sdae_ctr}
\vskip -0.2in
\end{figure}

\subsection{Autoencoders}\label{sec:ae}
An autoencoder (AE) is a feedforward neural network to encode the input into a more compact representation and reconstruct the input with the learned representation. In its simplest form, an autoencoder is no more than a multilayer perceptron with a bottleneck layer (a layer with a small number of hidden units) in the middle. The idea of autoencoders has been around for decades \cite{lecun-87,bourlard1988auto,dlbook} and abundant variants of autoencoders have been proposed to enhance representation learning including sparse AE \cite{poultney2006efficient}, contractive AE \cite{rifai2011contractive}, and denoising AE \cite{DBLP:journals/jmlr/VincentLLBM10}. For more details, please refer to a nice recent book on deep learning \cite{dlbook}. Here we introduce a kind of multilayer denoising AE, known as stacked denoising autoencoders (SDAE), both as an example of AE variants and as background for its applications on BDL-based recommender systems in Section \ref{sec:bdl}.

SDAE \cite{DBLP:journals/jmlr/VincentLLBM10} is a feedforward neural network for learning representations (encoding) of the input data by learning to predict the clean input itself in the output, as shown in Figure \ref{fig:sdae_ctr}. The hidden layer in the middle, i.e., $\X_2$ in the figure, can be constrained to be a bottleneck to learn compact representations. The difference between traditional AE and SDAE is that the input layer $\X_0$ is a \emph{corrupted} version of the \emph{clean} input data. Essentially an SDAE solves the following optimization problem:
%\footnote{*** Use $\|$ instead of $||$ for the norm. (** Modified.)}
\begin{align*}
\min\limits_{\{\W_l\},\{\b_l\}}& \|\X_c-\X_L\|_F^2+\lambda\sum\limits_l \|\W_l\|_F^2\\
\mbox{subject to}~~&\X_{l}=\sigma(\X_{l-1}\W_l+\b_l), l=1,\dots,L-1\\
&\X_{L}=\X_{L-1}\W_L+\b_L,
\end{align*}
Here SDAE can be regarded as a multilayer perceptron for regression tasks described in the previous section. The input $\X_0$ of the MLP is the corrupted version of the data and the target $\Y$ is the clean version of the data $\X_c$. For example, $\X_c$ can be the raw data matrix, and we can randomly set $30\%$ of the entries in $\X_c$ to $0$ and get $\X_0$. In a nutshell, SDAE learns a neural network that takes the noisy data as input and recovers the clean data in the last layer. This is what `denoising' means. Normally, the output of the middle layer, i.e., $\X_2$ in Figure \ref{fig:sdae_ctr}, would be used to compactly represent the data.

\begin{figure*}[!tb]
\begin{center}
\includegraphics[height=1.5cm]{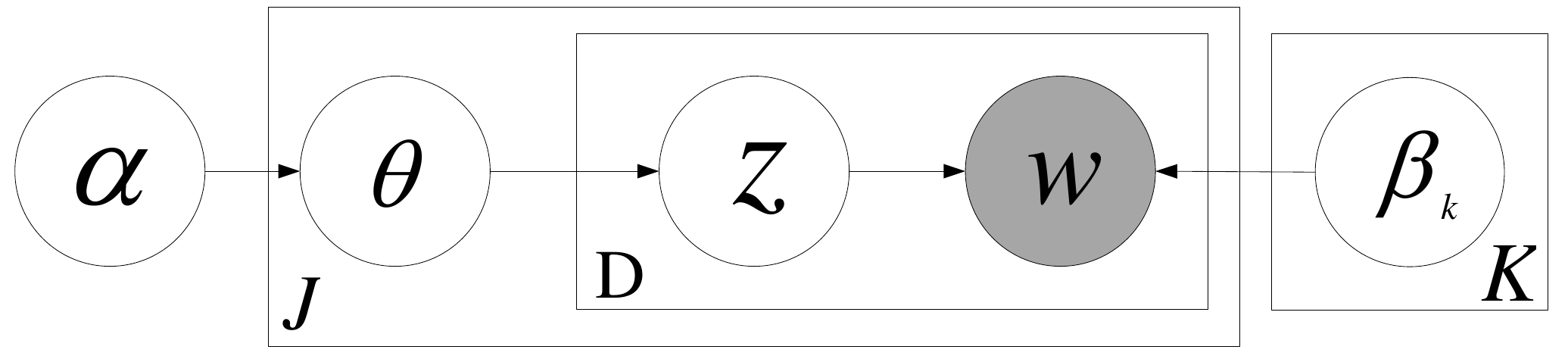}
\end{center}
\vskip -0.2in
\caption{The probabilistic graphical model for LDA, $J$ is the number of documents and $D$ is the number of words in a document.
}
\label{fig:lda}
\vskip -0.1in
\end{figure*}

\subsection{Other Deep Learning Models}
Other commonly used deep learning models include convolutional neural networks (CNN) \cite{hubel1968receptive,lecun1998gradient}, which apply convolution operators and pooling operators to process image or video data, and recurrent neural networks (RNN) \cite{hochreiter1997long,dlbook}, which use recurrent computation to imitate human memory, and restricted Boltzmann machines (RBM) \cite{RBM}, which are undirected probabilistic neural networks with binary hidden and visible layers. Note that there is a vast literature on deep learning and neural networks. The introduction in this section intends to serve only as the background of BDL. Readers are referred to \cite{dlbook} for a comprehensive survey and more details.
%Note that besides the deep learning models discussed above, there are many others, including restricted Boltzmann machine \cite{RBM}, sigmoid belief networks \cite{neal1992}, deep belief networks \cite{DBN}, and so on. Readers are referred to \cite{dlbook} for a comprehensive survey and more details.

\section{Probabilistic Graphical Models}\label{sec:pgm} %tocite
Probabilistic Graphical Models (PGM) use diagrammatic representations to describe random variables and relationships among them. Similar to a graph that contains nodes (vertices) and links (edges), PGM has nodes to represent random variables and links to express probabilistic relationships among them.

\subsection{Models}\label{sec:pgm_models}
As pointed out in \cite{PRML}, there are two main types of PGMs, directed PGMs (also known as Bayesian networks) and undirected PGMs (also known as Markov random fields), although there exist hybrid ones. In this paper we mainly focus on directed PGMs\footnote{For convenience, PGM stands for directed PGM in this paper unless specified otherwise.}. For details on undirected PGMs, readers are referred to \cite{PRML}.

A classic example of a PGM would be latent Dirichlet allocation (LDA), which is used as a topic model to analyze the generation of words and topics in documents. Usually PGM comes with a graphical representation of the model and a generative process to depict the story of how the random variables are generated step by step. Figure \ref{fig:lda} shows the graphical model for LDA and the corresponding generative process is as follows:
\begin{itemize}
\item For each document $j$ ($j=1,2,\dots,J$),
\begin{enumerate}
\item Draw topic proportions $\theta_j\sim\mbox{Dirichlet}(\alpha)$.
\item For each word $w_{jn}$ of item~(paper) $\w_j$,
\begin{enumerate}
\item Draw topic assignment $z_{jn}\sim\mbox{Mult}(\theta_j)$.
\item Draw word $w_{jn}\sim\mbox{Mult}(\beta_{z_{jn}})$.
\end{enumerate}
\end{enumerate}
\end{itemize}

The generative process above gives the story of how the random variables are generated. In the graphical model in Figure \ref{fig:lda}, the shaded node denotes observed variables while the others are latent variables ($\tha$ and $\z$) or parameters ($\alpha$ and $\beta$). As we can see, once the model is defined, learning algorithms can be applied to automatically learn the latent variables and parameters.

Due to its Bayesian nature, PGM like LDA is easy to extend to incorporate other information or to perform other tasks. For example, after LDA, different variants of topic models based on it have been proposed. The authors in \cite{DTM,cDTM} proposed to incorporate temporal information and \cite{CTM} extends LDA by assuming correlations among topics. To make it possible to process large datasets, \cite{onlineLDA} extends LDA from the batch mode to the online setting. On recommender systems, \cite{CTR} extends LDA to incorporate rating information and make recommendations. This model is then further extended to incorporate social information \cite{CTR-SMF,CTRSR,RCTR}.

\begin{table*}[!t]
\newcommand{\tabincell}[2]{\begin{tabular}{@{}#1@{}}#2\end{tabular}}
\caption{\small Summary of BDL Models. $\Om_h$ is the set of hinge variables mentioned in Section \ref{sec:general}. $\V$ and $\U$ are the item latent matrix and the user latent matrix~(Section \ref{sec:cdl}). $\S$ is the relational latent matrix (Section \ref{sec:rsdae}), and $\X$ is the content matrix (Section \ref{sec:dpfa_sbn}).
}\label{table:summary}
\begin{center}
\vskip -0.5cm
\begin{scriptsize}
\begin{tabular}{|c|l|c|l|c|c|c|}
\hline
 Applications & Models & $\Om_h$ & Variance of $\Om_h$ & MAP &  Gibbs Sampling & SG Thermostats \tabularnewline
\hline
\multirow{5}{*}{\tabincell{l}{Recommender\\ ~~~~Systems}} & CDL & $\{\V\}$ & Hyper-Variance & \checkmark & & \tabularnewline
\cline{2-7}
 & Bayesian CDL & $\{\V\}$ & Hyper-Variance & & \checkmark &  \tabularnewline
\cline{2-7}
& Marginalized CDL & $\{\V\}$ & Learnable Variance & \checkmark & &  \tabularnewline
\cline{2-7}
& Symmetric CDL & $\{\V,\U\}$ & Learnable Variance & \checkmark &  &  \tabularnewline
\cline{2-7}
& Collaborative Deep Ranking & $\{\V\}$ & Hyper-Variance & \checkmark &  &  \tabularnewline
\hline

\multirow{3}{*}{\tabincell{l}{~Topic\\Models}} & Relational SDAE & $\{\S\}$ & Hyper-Variance & \checkmark  & & \tabularnewline
\cline{2-7}
 & DPFA-SBN & $\{\X\}$ & Zero-Variance & & \checkmark & \checkmark \tabularnewline
\cline{2-7}
& DPFA-RBM & $\{\X\}$ & Zero-Variance & & \checkmark & \checkmark \tabularnewline
\hline

%\multirow{1}{*}{\tabincell{l}{Control}} & Embed to Control & $\{\z_t,\z_{t+1}\}$ & Learnable Variance &   & \checkmark & & \tabularnewline

%\hline
\end{tabular}
%\end{sc}
\end{scriptsize}
\end{center}
\vskip -0.2in
\end{table*}

\subsection{Inference and Learning}
Strictly speaking, the process of finding the parameters (e.g., $\alpha$ and $\beta$ in Figure \ref{fig:lda}) is called learning and the process of finding the latent variables (e.g., $\tha$ and $\z$ in Figure \ref{fig:lda}) given the parameters is called inference. However, given only the observed variables (e.g., $\w$ in Figure \ref{fig:lda}), learning and inference are often intertwined. Usually, the learning and inference of LDA would alternate between the updates of latent variables (which correspond to inference) and the updates of the parameters (which correspond to learning). Once the learning and inference of LDA is completed, we would have the parameters $\alpha$ and $\beta$. If a new document arrives, we can now fix the learned $\alpha$ and $\beta$ and then perform inference alone to find the topic proportions $\theta_j$ of the new document.\footnote{For convenience, we use `learning' to represent both `learning and inference' in the following text.}

As in LDA, various learning and inference algorithms are available for each PGM. Among them, the most cost-effective one is probably maximum a posteriori (MAP), which amounts to maximizing the posterior probability of the latent variable. Using MAP, the learning process is equivalent to minimizing (or maximizing) an objective function with regularization. One famous example is the probabilistic matrix factorization (PMF) \cite{PMF}. The learning of the graphical model in PMF is equivalent to the factorization of a large matrix into two low-rank matrices with L2 regularization.

MAP, as efficient as it is, gives us only \emph{point estimates} of latent variables (and parameters). In order to take the uncertainty into account and harness the full power of Bayesian models, one would have to resort to Bayesian treatments such as variational inference and Markov chain Monte Carlo (MCMC). For example, the original LDA uses variational inference to approximate the true posterior with factorized variational distributions \cite{LDA}. Learning of the latent variables and parameters then boils down to minimizing the KL-divergence between the variational distributions and the true posterior distributions. Besides variational inference, another choice for a Bayesian treatment is to use MCMC. For example, MCMC algorithms such as \cite{porteous2008fast} have been proposed to learn the posterior distributions of LDA.

\section{Bayesian Deep Learning}\label{sec:bdl}
With the background on deep learning and PGM, we are now ready to introduce the general framework and some concrete examples of BDL. Specifically, in this section we will list some recent BDL models with applications on recommender systems and topic models. A summary of these models is shown in Table \ref{table:summary}.

\subsection{General Framework}\label{sec:general}
As mentioned in Section \ref{sec:intro}, BDL is a principled probabilistic framework with two seamlessly integrated components: a \emph{perception component} and a \emph{task-specific component}.

\begin{figure}[!tb]
\begin{center}
\includegraphics[height=2.0cm]{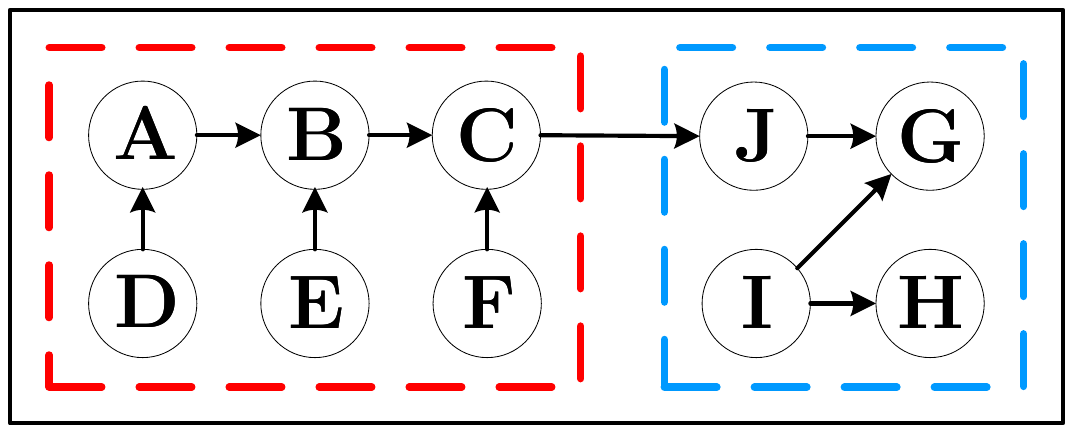}
\end{center}
\vskip -0.2in
\caption{The PGM for an example BDL. The red rectangle on the left indicates the perception component, and the blue rectangle on the right indicates the task-specific component. The hinge variable $\Om_h=\{\J\}$.
}
\label{fig:BDL_typeI}
\vskip -0.2in
\end{figure}

\textbf{PGM for BDL}: Figure \ref{fig:BDL_typeI} shows the PGM of a simple BDL model as an example. The part inside the red rectangle on the left represents the perception component and the part inside the blue rectangle on the right is the task-specific component. Typically, the perception component would be a probabilistic formulation of a deep learning model with multiple nonlinear processing layers represented as a chain structure in the PGM. While the nodes and edges in the perception component are relatively simple, those in the task-specific component often describe more complex distributions and relationships among variables (as in LDA).

\textbf{Three Sets of Variables}: There are three sets of variables in a BDL model: perception variables, hinge variables, and task variables: (1) In this paper, we use $\Om_p$ to denote the set of perception variables (e.g., $\A$, $\B$, and $\C$ in Figure \ref{fig:BDL_typeI}), which are the variables in the perception component. Usually $\Om_p$ would include the weights and neurons in the probabilistic formulation of a deep learning model. (2) We use $\Om_h$ to denote the set of hinge variables (e.g. $\J$ in Figure \ref{fig:BDL_typeI}). These variables directly interact with the perception component from the task-specific component. Table \ref{table:summary} shows the set of hinge variables $\Om_h$ for each listed BDL models. (3) The set of task variables (e.g. $\G$, $\I$, and $\H$ in Figure \ref{fig:BDL_typeI}), i.e., variables in the task-specific component without direct relation to the perception component, is denoted as $\Om_t$.

\textbf{The I.I.D. Requirement}: Note that hinge variables are always in the task-specific component. Normally, the connections between hinge variables $\Om_h$ and the perception component (e.g., $\C\rightarrow \J$ in Figure \ref{fig:BDL_typeI}) should be i.i.d. for convenience of parallel computation in the perception component. For example, each row in $\J$ is related to only one corresponding row in $\C$. Although it is not mandatory in BDL models, meeting this requirement would significantly increase the efficiency of parallel computation in model training.

\textbf{Joint Distribution Decomposition}: If the edges between the two components \emph{point towards} $\Om_h$ (as shown in Figure~ \ref{fig:BDL_typeI}, where $\Om_p=\{\A, \B, \C, \D, \E, \F\}$, $\Om_h=\{\J\}$, and $\Om_t=\{\I, \G, \H\}$), the joint distribution of all variables can be written as:
\begin{align}
p(\Om_p,\Om_h,\Om_t)=p(\Om_p)p(\Om_h|\Om_p)p(\Om_t|\Om_h). \label{eq:typeone}
\end{align}

If the edges between the two components \emph{originate from} $\Om_h$ (similar to Figure \ref{fig:BDL_typeI} except that the edge points from $\J$ to $\C$), the joint distribution of all variables can be written as:
\begin{align}
p(\Om_p,\Om_h,\Om_t)=p(\Om_t)p(\Om_h|\Om_t)p(\Om_p|\Om_h). \label{eq:typetwo}
\end{align}

Apparently, it is possible for BDL to have some edges between the two components pointing towards $\Om_h$ and some originating from $\Om_h$, in which case the decomposition of the joint distribution would be more complex.

\textbf{Variance Related to $\Om_h$}: As mentioned in Section \ref{sec:intro}, one of the motivations for BDL is to model the \emph{uncertainty of exchanging information} between the perception component and the task-specific component, which boils down to modeling the uncertainty related to $\Om_h$. For example, this kind of uncertainty is reflected in the variance of the conditional density $p(\Om_h|\Om_p)$ in Equation (\ref{eq:typeone})\footnote{For models with the joint likelihood decomposed as in Equation~(\ref{eq:typetwo}), the uncertainty is reflected in the variance of $p(\Om_p|\Om_h)$.}. According to the degree of flexibility, there are three types of variance for $\Om_h$ (for simplicity we assume the joint likelihood of BDL is Equation (\ref{eq:typeone}), $\Om_p=\{p\}$, $\Om_h=\{h\}$, and $p(\Om_h|\Om_p)=\NM(h|p,s)$ in our example):
\begin{compactitem}
\item \textbf{Zero-Variance}: Zero-Variance (ZV) assumes no uncertainty during the information exchange between the two components. In the example, zero-variance means directly setting $s$ to $0$.
\item \textbf{Hyper-Variance}: Hyper-Variance (HV) assumes that uncertainty during the information exchange is defined through hyperparameters. In the example, HV means that $s$ is a hyperparameter that is manually tuned.
\item \textbf{Learnable Variance}: Learnable Variance (LV) uses learnable parameters to represent uncertainty during the information exchange. In the example, $s$ is the learnable parameter.
\end{compactitem}
As shown above, we can see that in terms of model flexibility, $\text{LV}>\text{HV}>\text{ZV}$. Normally, if the models are properly regularized, an LV model would outperform an HV model, which is superior to a ZV model. In Table \ref{table:summary}, we show the types of variance for $\Om_h$ in different BDL models. Note that although each model in the table has a specific type, one can always adjust the models to devise their counterparts of other types. For example, while CDL in the table is an HV model, we can easily adjust $p(\Om_h|\Om_p)$ in CDL to devise its ZV and LV counterparts. In \cite{CDL}, authors compare the performance of an HV CDL and a ZV CDL and finds that the former performs significantly better, meaning that sophisticatedly modeling uncertainty between two components is essential for performance.

\textbf{Learning Algorithms}: Due to the nature of BDL, practical learning algorithms need to meet these criteria:
\begin{compactenum}
\item They should be online algorithms in order to scale well for large datasets.
\item They should be efficient enough to scale linearly with the number of free parameters in the perception component.
\end{compactenum}
Criterion (1) implies that conventional variational inference or MCMC methods are not applicable. Usually an online version of them is needed \cite{onlineVB}. Most SGD-based methods do not work either unless only MAP inference (as opposed to Bayesian treatments) is performed. Criterion (2) is needed because there are typically a large number of free parameters in the perception component. This means methods based on Laplace approximation \cite{mackay1992practical} are not realistic since they involve the computation of a Hessian matrix that scales quadratically with the number of free parameters.

%\begin{figure}[!tb]
%\begin{center}
%\includegraphics[height=2.0cm]{fig/BDL_typeII}
%\end{center}
%\vskip -0.2in
%\caption{PGM for type II BDL.
%}
%\label{fig:BDL_typeII}
%\vskip -0.1in
%\end{figure}
%
%\begin{figure}[!tb]
%\begin{center}
%\includegraphics[height=2.0cm]{fig/BDL_typeIII}
%\end{center}
%\vskip -0.2in
%\caption{PGM for type III BDL.
%}
%\label{fig:BDL_typeIII}
%\vskip -0.2in
%\end{figure}

\subsection{Bayesian Deep Learning for Recommender Systems}\label{sec:recsys}
Despite the successful applications of deep learning on natural language processing and computer vision, very few attempts have been made to develop deep learning models for CF. The authors in \cite{DBLP:conf/icml/SalakhutdinovMH07} use restricted Boltzmann machines instead of the conventional matrix factorization formulation to perform CF and \cite{DBLP:conf/icml/GeorgievN13} extends this work by incorporating user-user and item-item correlations. Although these methods involve both deep learning and CF, they actually belong to CF-based methods because they do not incorporate content information as in CTR \cite{CTR}, which is crucial for accurate recommendation. The authors in \cite{DBLP:conf/icassp/SainathKSAR13} use low-rank matrix factorization in the last weight layer of a deep network to significantly reduce the number of model parameters and speed up training, however it is for classification instead of recommendation tasks. On music recommendation, \cite{DBLP:conf/nips/OordDS13,DBLP:conf/mm/WangW14} directly use conventional CNN or deep belief networks (\mbox{DBN}) to assist representation learning for content information, but the deep learning components of their models are deterministic without modeling the noise and hence they are less robust. The models achieve performance boost mainly by loosely coupled methods without exploiting the interaction between content information and ratings. Besides, the CNN is linked directly to the rating matrix, which means the models will perform poorly due to serious overfitting when the ratings are sparse.

\subsubsection{Collaborative Deep Learning}\label{sec:cdl}
To address the challenges above, a hierarchical Bayesian model called collaborative deep learning (CDL) as a novel tightly coupled method for RS is introduced in \cite{CDL}. Based on a Bayesian formulation of SDAE, CDL tightly couples deep representation learning for the content information and collaborative filtering for the rating (feedback) matrix, allowing two-way interaction between the two.  Experiments show that CDL significantly outperforms the state of the art.

In the following text, we will start with the introduction of the notation used during our presentation of CDL. After that we will review the design and learning of CDL.

\textbf{Notation and Problem Formulation}:
Similar to the work in \cite{CTR}, the recommendation task considered in CDL takes implicit feedback \cite{DBLP:conf/icdm/HuKV08} as the training and test data.  The entire collection of $J$ items (articles or movies) is represented by a $J$-by-$B$ matrix $\X_c$, where row $j$ is the bag-of-words vector $\X_{c,j*}$ for item $j$ based on a vocabulary of size $B$.  With $I$ users, we define an $I$-by-$J$ binary rating matrix $\R=[\R_{ij}]_{I\times J}$. For example, in the dataset \emph{citeulike-a} \cite{CTR,CTRSR,CDL} $\R_{ij}=1$ if user $i$ has article $j$ in his or her personal library and $\R_{ij}=0$ otherwise. Given part of the ratings in $\R$ and the content information $\X_c$, the problem is to predict the other ratings in $\R$. Note that although CDL in its current form focuses on movie recommendation (where plots of movies are considered as content information) and article recommendation like \cite{CTR} in this section, it is general enough to handle other recommendation tasks (e.g., tag recommendation).

Matrix $\X_c$ plays the role of clean input to the SDAE while the noise-corrupted matrix, also a $J$-by-$B$ matrix, is denoted by $\X_0$.  The output of layer $l$ of the SDAE is denoted by $\X_l$ which is a $J$-by-$K_l$ matrix, where $K_l$ is the number of units in layer $l$. Similar to $\X_c$, row $j$ of $\X_l$ is denoted by $\X_{l,j*}$.  $\W_l$ and $\b_l$ are the weight matrix and bias vector, respectively, of layer $l$, $\W_{l,*n}$ denotes column $n$ of $\W_l$, and $L$ is the number of layers. For convenience, we use $\W^+$ to denote the collection of all layers of weight matrices and biases. Note that an $L/2$-layer SDAE corresponds to an $L$-layer network.

\begin{figure*}[!tb]
\begin{center}
\subfigure{
\includegraphics[height=3.6cm]{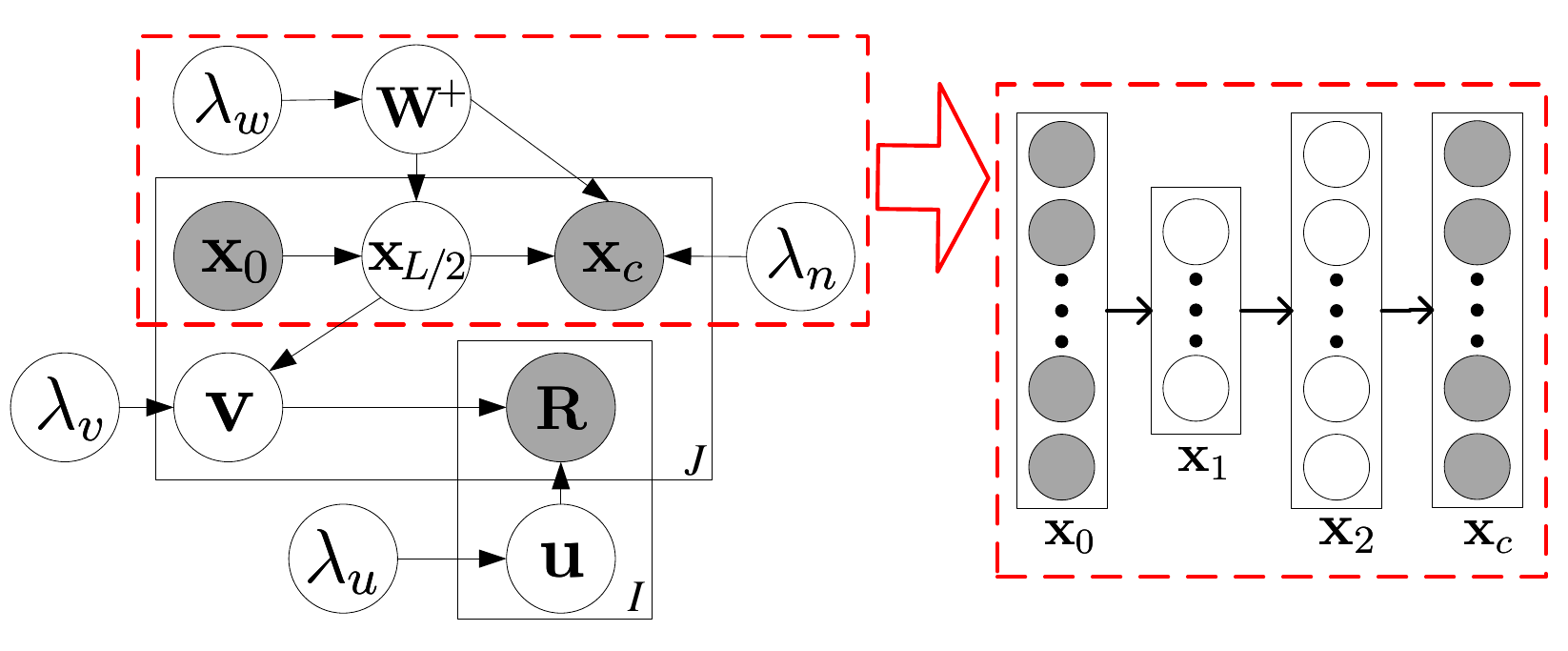}}
\hspace{0.3in}
\subfigure{
\includegraphics[height=3.4cm]{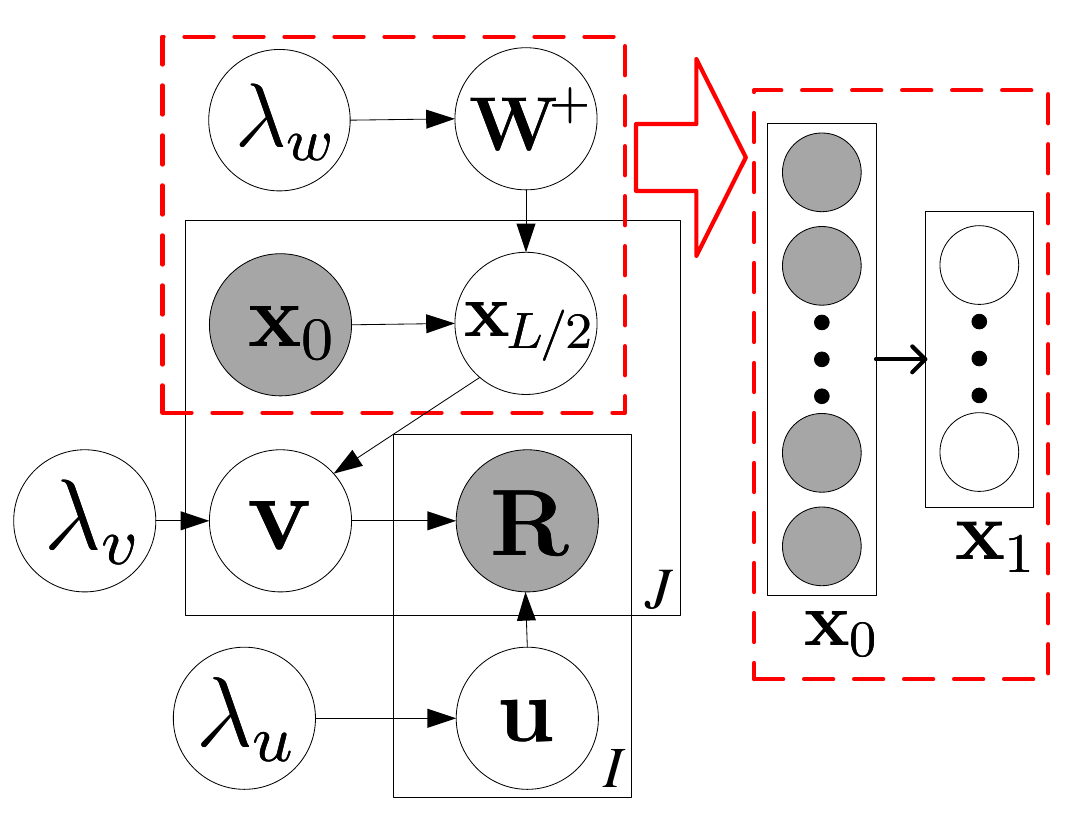}}
\end{center}
\vskip -0.2in
\caption{On the left is the graphical model of CDL. The part inside the dashed rectangle represents an SDAE.  An example SDAE with $L = 2$ is shown. On the right is the graphical model of the degenerated CDL. The part inside the dashed rectangle represents the encoder of an SDAE. An example SDAE with $L=2$ is shown on its right. Note that although $L$ is still $2$, the decoder of the SDAE vanishes. To prevent clutter, we omit all variables $\x_l$ except $\x_0$ and $\x_{L/2}$ in the graphical models.
}
\label{fig:cdl_pgm}
\vskip -0.2in
\end{figure*}

\textbf{Generalized Bayesian SDAE}:
Following the introduction of SDAE in Section \ref{sec:ae}, if we assume that both the clean input $\X_c$ and the corrupted input $\X_0$ are observed, similar to \cite{PRML,mackay1992practical,DBLP:conf/nips/BengioYAV13,nmSDAE},
%\footnote{*** They don't define a generative process, do they? (** References added and generative process modified.)}
we can define the following generative process of generalized Bayesian SDAE:
\begin{enumerate}
\item For each layer $l$ of the SDAE network,
\begin{enumerate}
\item For each column $n$ of the weight matrix $\W_l$, draw
\begin{align*}
\W_{l,*n} \sim \NM(\0,\lambda_w^{-1} \I_{K_l}).
\end{align*}
\item Draw the bias vector $\b_l \sim \NM(\0,\lambda_w^{-1} \I_{K_l})$.
\item For each row $j$ of $\X_l$, draw
\begin{align}\label{eq:gaussian}
\X_{l,j*} \sim \NM(\sigma(\X_{l-1,j*}\W_l+\b_l),\lambda_s^{-1} \I_{K_l}).
%\X_{l,j*} \sim \mbox{Delta}(\sigma(\X_{l-1,j*}\W_l+\b_l)).
\end{align}
%\item Generate the output of layer $l$, $\X_l = \sigma(\X_{l-1}\W_l+\b_l)$.
\end{enumerate}
\item For each item $j$, draw a clean input
\footnote{Note that while generation of the \emph{clean} input $\X_c$ from $\X_L$ is part of the generative process of the Bayesian SDAE, generation of the \emph{noise-corrupted} input $\X_0$ from $\X_c$ is an artificial noise injection process to help the SDAE learn a more robust feature representation.}
\begin{align*}
\X_{c,j*} \sim \NM(\X_{L,j*},\lambda_n^{-1}\I_{B}).
\end{align*}
%\footnote{*** So you assume that it has unit variance? (** Yes.) (*** Why isn't $\lambda_n$ introduced here, but later?) (** It might be better to introduce it here to keep it consistent with the latter. Delta distribution is used so that $\X$ can be treated as `generalized' random variables drawn from some distribution.)}
\end{enumerate}
Note that if $\lambda_s$ goes to infinity, the Gaussian distribution in Equation (\ref{eq:gaussian}) will become a Dirac delta distribution \cite{strichartz2003guide} centered at $\sigma(\X_{l-1,j*}\W_l+\b_l)$, where $\sigma(\cdot)$ is the sigmoid function. The model will degenerate to be a Bayesian formulation of SDAE. That is why we call it generalized SDAE.
%We note that it is equivalent to a Gaussian distribution
%\begin{align*}
%\NM(\sigma(\X_{l-1,j*}\W_l+\b_l),\lambda^{-1} \I_{K_l})
%\end{align*}
%with $\lambda$ approaching infinity.

%\footnote{** Dirac delta distribution is the core of real analysis and is one of the most important building block in signal processing and physics. That means there is a deep background for using Dirac delta function. Should we mention this in the paper? (*** I don't think you have space for it.)}

%where $J$ is the number of items, $L$ is the number of layers, $\W_{l,*n}$ means the n'th column of weight matrix $\W_l$, and $\X_{c,j*}$ is the j'th row of the clean input matrix $\X_c$.
%\footnote{*** Is it necessary to mention this again as you have already introduced it in Section 2? (** Deleted redundant part, and moved non-redundant part to Section 2.)}

Note that the first $L/2$ layers of the network act as an encoder and the last $L/2$ layers act as a decoder.  Maximization of the posterior probability is equivalent to minimization of the reconstruction error with weight decay taken into consideration.
%Note that practice the weight decay of $\b_l$ is $0$, which makes the step 1(c) unnecessary.
%\footnote{*** This sentence may not be very clear to readers. (** Deleted. Actually, in the my code deepmat and deeplearntoolbox, weight decay are also used for biases, which seem to work well. To avoid confusion, I guess we'd better delete this sentence and use weigh decay also for biases in our models.)}

\textbf{Collaborative Deep Learning}: Using the Bayesian SDAE as a component, the generative process of CDL is defined as follows:
\begin{compactenum}
\item Generate variables of generalized Bayesian SDAE.
\item For each item $j$,
\begin{compactenum}
\item Draw the latent item offset vector $\ep_j \sim \NM(\0,\lambda_v^{-1}\I_K)$ and then set the latent item vector:
$
\v_j=\ep_j+\X_{\frac{L}{2},j*}^T.
$
\end{compactenum}
\item\label{enum:user} Draw a latent user vector for each user $i$:
$$
\u_i \sim \NM(\0,\lambda_u^{-1}\I_K).
$$
\item\label{enum:rating} Draw a rating $\R_{ij}$ for each user-item pair $(i,j)$:
$
\R_{ij} \sim \NM(\u_i^T\v_j,\C_{ij}^{-1}).\\
$
\end{compactenum}
Here $\lambda_w$, $\lambda_n$, $\lambda_u$, $\lambda_s$, and $\lambda_v$ are hyperparameters
%\footnote{*** Why is $\lambda_n$ mentioned specifically but not the other hyperparameters? (** Modified.)}
and $\C_{ij}$ is a confidence parameter similar to that for CTR \cite{CTR} ($\C_{ij} = a$ if $\R_{ij}=1$ and $\C_{ij}=b$ otherwise). Note that the middle layer $\X_{L/2}$ serves as a bridge between the ratings and content information. This middle layer, along with the latent offset $\ep_j$, is the key that enables CDL to simultaneously learn an effective feature representation and capture the similarity and (implicit) relationship between items (and users). Similar to the generalized SDAE, for computational efficiency, we can also take $\lambda_s$ to infinity.

The graphical model of CDL when $\lambda_s$ approaches positive infinity is shown in Figure \ref{fig:cdl_pgm}, where, for notational simplicity, we use $\x_0$, $\x_{L/2}$, and $\x_C$ in place of $\X_{0,j*}^T$, $\X_{\frac{L}{2},j*}^T$, and $\X_{c,j*}^T$, respectively.

Note that according the definition in Section \ref{sec:general}, here the perception variables $\Om_p=\{\{\W_l\},\{\b_l\},\{\X_l\},\X_c\}$, the hinge variables $\Om_h=\{\V\}$, and the task variables $\Om_t=\{\U,\R\}$, where $\V=(\v_j)_{j=1}^J$ and $\U=(\u_i)_{i=1}^I$.
%$\W^+$ is a shorthand for weights $\W_l$ and biases $\b_l$ for all $L$ layers.
%\footnote{*** Already introduced in Section 2. (** Deleted.)}
%\footnote{*** The last two sentences are redundant.  Either keep them here or in the figure capture, but not both. (** Deleted.)}

%\begin{figure} [tb]
%\begin{center}
%  \begin{tabular}{ccc}
%   \includegraphics*[height=35mm]{fig/cdl_with_sdae}
%  \end{tabular} %\vskip -0.4cm
%\caption{\small Graphical model of CDL. The part inside the dashed rectangle represents an SDAE.  A sample SDAE with $L = 2$ is shown on the right-hand side.}
%\label{fig:cdl_graphmodel}
%\end{center}
%\vskip -0.5cm
%\end{figure}
\textbf{Learning}:
Based on the CDL model above, all parameters could be treated as random variables so that fully Bayesian methods such as Markov chain Monte Carlo (MCMC) or variational approximation methods \cite{DBLP:journals/ml/JordanGJS99} may be applied.  However, such treatment typically incurs high computational cost. Consequently, CDL uses an EM-style algorithm for obtaining the MAP estimates, as in \cite{CTR}.

As in CTR \cite{CTR}, maximizing the posterior probability is equivalent to maximizing the joint log-likelihood of $\U$, $\V$, $\{\X_l\}$, $\X_c$, $\{\W_l\}$, $\{\b_l\}$, and $\R$ given $\lambda_u$, $\lambda_v$, $\lambda_w$, $\lambda_s$, and $\lambda_n$:
\begin{align}
\mathscr{L}=&-\frac{\lambda_u}{2}\sum\limits_i \|\u_i\|_2^2
-\frac{\lambda_w}{2}\sum\limits_l(\|\W_l\|_F^2+\|\b_l\|_2^2) \nonumber\\
&-\frac{\lambda_v}{2}\sum\limits_j\|\v_j-\X_{\frac{L}{2},j*}^T\|_2^2-\frac{\lambda_n}{2}\sum\limits_j\|\X_{L,j*}-\X_{c,j*}\|_2^2 \nonumber\\ &-\frac{\lambda_s}{2}\sum\limits_l\sum\limits_j\|\sigma(\X_{l-1,j*}\W_l+\b_l)-\X_{l,j*}\|_2^2 \nonumber \\
&-\sum\limits_{i,j}\frac{\C_{ij}}{2}(\R_{ij}-\u_i^T\v_j)^2. \nonumber
\end{align}\label{eq:Lgen}
If $\lambda_s$ goes to infinity, the likelihood becomes:
%\footnote{*** Better use $\{\X_l\}, \{\W_l\}, \{\b_l\}$. (** Modified. I also added the indexes but don't know if it's necessary. Please delete them if it is not since it looks kind of messy with the indexes.)}
%\begin{align}\label{eq:L}
%\mathscr{L}=&-\frac{\lambda_u}{2}\sum\limits_i \u_i^T\u_i
%-\frac{\lambda_w}{2}\sum\limits_l(\|\W_l\|_F^2+\|\b_l\|_2^2) \nonumber \\
%&-\frac{\lambda_v}{2}\sum\limits_j(\v_j-f_e(\X_{0,j*},\W^+))^T(\v_j-f_e(\X_{0,j*},\W^+)) \nonumber \\
%&-\frac{\lambda_n}{2}\sum\limits_j(f_r(\X_{0,j*},\W^+)-\X_{c,j*})^T(f_r(\X_{0,j*},\W^+)-\X_{c,j*})
%-\sum\limits_{i,j}\frac{c_{ij}}{2}(r_{ij}-\u_i^T\v_j)^2,
%\nonumber
%\end{align}
\begin{align}\label{eq:L}
\mathscr{L}=&-\frac{\lambda_u}{2}\sum\limits_i \|\u_i\|_2^2
-\frac{\lambda_w}{2}\sum\limits_l(\|\W_l\|_F^2+\|\b_l\|_2^2)\nonumber \\
&-\frac{\lambda_v}{2}\sum\limits_j\|\v_j-f_e(\X_{0,j*},\W^+)^T\|_2^2 \nonumber \\
&-\frac{\lambda_n}{2}\sum\limits_j\|f_r(\X_{0,j*},\W^+)-\X_{c,j*}\|_2^2 \nonumber \\
&-\sum\limits_{i,j}\frac{\C_{ij}}{2}(\R_{ij}-\u_i^T\v_j)^2,
\end{align}
where the encoder function $f_e(\cdot,\W^+)$ takes the corrupted content vector $\X_{0,j*}$ of item $j$ as input and computes the encoding of the item, and the function $f_r(\cdot,\W^+)$ also takes $\X_{0,j*}$ as input, computes the encoding and then reconstructs the content vector of item $j$.
%\footnote{*** The name reconstruction function is not exactly correct because it actually performs \emph{both} encoding and reconstruction, taking $\X_{0,j*}$ as input. (** Modified. But don't know if it is clear enough.)}
For example, if the number of layers $L=6$, $f_e(\X_{0,j*},\W^+)$ is the output of the third layer while $f_r(\X_{0,j*},\W^+)$ is the output of the sixth layer.

\begin{figure}[!tb]
\begin{center}
%\framebox[4.0in]{$\;$}
%\includegraphics[height=5cm]{likeli1.eps}
\vskip -0.23in
\includegraphics[height=6.0cm]{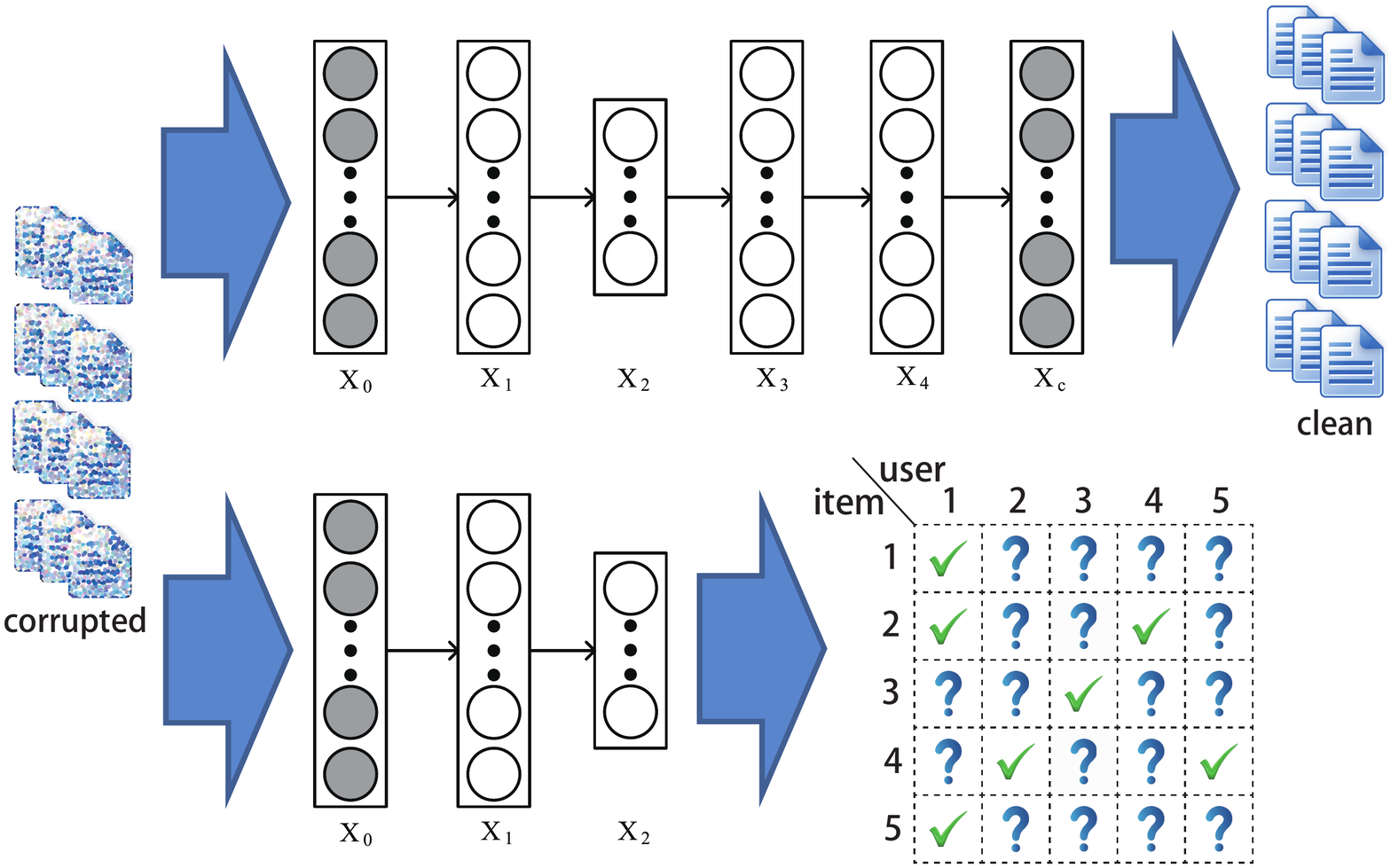}
\end{center}
\vskip -0.4in
\caption{NN representation for degenerated CDL.
}
\label{fig:twonet}
\vskip -0.2in
\end{figure}

%\begin{align*}
%\mathscr{L}=&-\frac{\lambda_u}{2}\sum\limits_i \|\u_i\|_2^2
%-\frac{\lambda_v}{2}\sum\limits_j\|\v_j-f_e(\X_{0,j*},\W^+)^T\|_2^2  \\
%&-\frac{\lambda_n}{2}\sum\limits_j\|f_r(\X_{0,j*},\W^+)-\X_{c,j*}\|_2^2 \\
%&-\sum\limits_{i,j}\frac{\C_{ij}}{2}(\R_{ij}-\u_i^T\v_j)^2,
%\end{align*}

From the optimization perspective, the third term in the objective function (\ref{eq:L}) above is equivalent to a multi-layer perceptron using the latent item vectors $\v_j$ as the target while the fourth term is equivalent to an SDAE minimizing the reconstruction error. From the perspective of neural networks (NN), when $\lambda_s$ approaches positive infinity, training of the probabilistic graphical model of CDL in Figure \ref{fig:cdl_pgm}(left) would degenerate to simultaneously training two neural networks overlaid together with a common input layer (the corrupted input) but different output layers, as shown in Figure \ref{fig:twonet}. Note that the second network is much more complex than typical neural networks due to the involvement of the rating matrix.
%\footnote{Note that empirically the bias $\b_l$ does not need to have regularizer (corresponding to weight decay in neural network training). Since in our experiments the existence of weight decay term for bias $\b_l$ has little impact on the performance, we will keep the term here.}
%\footnote{*** I find the remark in the previous footnote is a bit strange.  Since weight decay for the bias has little impact, why is it still kept in the equation? (** Footnote deleted. Since in the code we also use weight decay on biases, which works just as well. I guess we will skip the footnote to avoid confusion.)}

When the ratio $\lambda_n/\lambda_v$ approaches positive infinity, it will degenerate to a two-step model in which the latent representation learned using SDAE is put directly into the CTR. The interaction between the perception component and the task-specific component is one-way (from the perception component to the task-specific component), meaning that the perception component will not be affected by the task-specific component. Another extreme happens when $\lambda_n/\lambda_v$ goes to zero where the decoder of the SDAE essentially vanishes.  On the right of Figure \ref{fig:cdl_pgm} is the graphical model of the degenerated CDL when $\lambda_n/\lambda_v$ goes to zero.
%\footnote{*** Again these two sentences are redundant. (** Deleted.)}
As demonstrated in the experiments, the predictive performance will suffer greatly for both extreme cases \cite{CDL}. This verifies that (1) the information from the task-specific component can improve the perception component, and (2) mutual boosting effect is crucial to BDL.

%\begin{figure} [tb]
%\begin{center}
%  \begin{tabular}{ccc}
%   \includegraphics*[height=35mm]{fig/degenerated_cdl_with_sdae}
%  \end{tabular} %\vskip -0.4cm
%\caption{\small Graphical model of degenerated CDL. The part inside the dashed rectangle represents the encoder of an SDAE. A sample SDAE with $L=2$ is shown on the right-hand side.  Note that although $L$ is still $2$, the decoder of the SDAE vanishes.} \label{fig:pgm_cdl_encode}
%\end{center}
%\end{figure}
For $\u_i$ and $\v_j$, block coordinate descent similar to \cite{CTR,DBLP:conf/icdm/HuKV08} is used. Given the current $\W^+$, we compute the gradients of $\mathscr{L}$ with respect to $\u_i$ and $\v_j$ and then set them to zero, leading to the following update rules:
\begin{align}
\u_i&\leftarrow(\V \C_i \V^T+\lambda_u \I_K)^{-1}\V \C_i \R_i \nonumber \\
\v_j&\leftarrow(\U \C_i \U^T+\lambda_v \I_K)^{-1}(\U \C_j \R_j+\lambda_v f_e(\X_{0,j*},\W^+)^T), \nonumber
\end{align}
where $\U=(\u_i)^I_{i=1}$, $\V=(\v_j)^J_{j=1}$, $\C_i = \mbox{diag}(\C_{i1},\ldots,\C_{iJ})$ is a diagonal matrix,
%\footnote{*** $\C_i = \mbox{diag}(c_{i1},\ldots,c_{iJ})$ (** Modified.)}
$\R_i = (\R_{i1},\ldots,\R_{iJ})^T$ is a column vector containing all the ratings of user $i$,
%\footnote{*** $\R_i = (r_{i1},\ldots,r_{iJ})^T$ (** Modified.)}
and $\C_{ij}$ reflects the confidence controlled by $a$ and $b$ as discussed in \cite{DBLP:conf/icdm/HuKV08}. $\C_j$ and $\R_j$ are defined similarly for item $j$.

Given $\U$ and $\V$, we can learn the weights $\W_l$ and biases $\b_l$ for each layer using the back-propagation learning algorithm. The gradients of the likelihood with respect to $\W_l$ and $\b_l$ are as follows:
\begin{align*}
&\nabla_{\W_l}\mathscr{L} = -\lambda_w\W_l\\
&-\lambda_v\sum\limits_j\nabla_{\W_l}f_e(\X_{0,j*},\W^+)^T(f_e(\X_{0,j*},\W^+)^T-\v_j)\\
&-\lambda_n\sum\limits_j\nabla_{\W_l}f_r(\X_{0,j*},\W^+)(f_r(\X_{0,j*},\W^+)-\X_{c,j*})
\end{align*}
\begin{align*}
&\nabla_{\b_l}\mathscr{L} = -\lambda_w\b_l\\
&-\lambda_v\sum\limits_j\nabla_{\b_l}f_e(\X_{0,j*},\W^+)^T(f_e(\X_{0,j*},\W^+)^T-\v_j)\\
&-\lambda_n\sum\limits_j\nabla_{\b_l}f_r(\X_{0,j*},\W^+)(f_r(\X_{0,j*},\W^+)-\X_{c,j*}).
\end{align*}
By alternating the update of $\U$, $\V$, $\W_l$, and $\b_l$, we can find a local optimum for $\mathscr{L}$. Several commonly used techniques such as using a momentum term may be applied to alleviate the local optimum problem. Note that a carefully designed BDL model (according to the i.i.d. requirement and with proper variance models as stated in Section \ref{sec:general}) can minimize the overhead of seamlessly combining the perception component and the task-specific component. In CDL, the computational complexity (per iteration) of the perception component is $O(JBK_1)$ and that of the task-specific component is $O(K^2N_R+K^3)$, where $N_R$ is the number of non-zero entries in the rating matrix and $K=K_{\frac{L}{2}}$. The computational complexity (per iteration) for the whole model is $O(JBK_1+K^2N_R+K^3)$ \cite{CDL}. No significant overhead is introduced.

\begin{table}[!tb]
\begin{scriptsize}
\centering
\vskip -0.1cm
\caption{Recall@300 on the dataset \emph{citeulike-a} ($\%$)}\label{table:cdl_recall}
\vskip -0.2cm
\begin{tabular}{|c|c|c|c|c|} \hline
SVDFeature \cite{DBLP:journals/jmlr/ChenZLCZY12} & CMF \cite{DBLP:conf/kdd/SinghG08} & DeepMusic \cite{DBLP:conf/nips/OordDS13} & CTR \cite{CTR} & CDL \cite{CDL} \\ \hline
11.19 &13.45	&12.32&	22.11 & \textbf{31.06} \\ \hline
\end{tabular}
\vskip -0.5cm
\end{scriptsize}
\end{table}

\textbf{Prediction}: Let $D$ be the observed test data. Similar to \cite{CTR}, CDL uses the point estimates of $\u_i$, $\W^+$
%\footnote{** Typo fixed.}
and $\ep_j$ to calculate the predicted rating:
\begin{align}
E[\R_{ij}|D]\approx E[\u_i|D]^T(E[f_e(\X_{0,j*},\W^+)^T|D]+E[\ep_j|D]),\nonumber
\end{align}
where $E[\cdot]$ denotes the expectation operation.
In other words, we approximate the predicted rating as:%\footnote{*** Note that weird hyphenation problem: word-s.}
%\footnote{*** You haven't defined the terms in-matrix prediction and out-of-matrix prediction (even though they are used in the CTR paper). (** Modified. Since we performs in-matrix and out-of-matrix prediction together, there might not be necessary to treat them separately.)}
\begin{align}
\R^*_{ij}\approx(\u^*_j)^T(f_e(\X_{0,j*},{\W^+}^*)^T+\ep^*_j)=(\u^*_i)^T\v^*_j .\nonumber
\end{align}
Note that for any new item $j$ with no rating in the training data, its offset $\ep^*_j$ will be $\0$.

Table \ref{table:cdl_recall} shows the recall of recommendation with $300$ recommended items for different methods in the dataset \emph{citeulike-a}. Please refer to \cite{CDL} for more details.

In the following text, we provide several extensions of CDL from different perspectives.

\begin{figure}[!tb]
\begin{center}
%\framebox[4.0in]{$\;$}
%\includegraphics[height=5cm]{likeli1.eps}
\includegraphics[height=4.0cm]{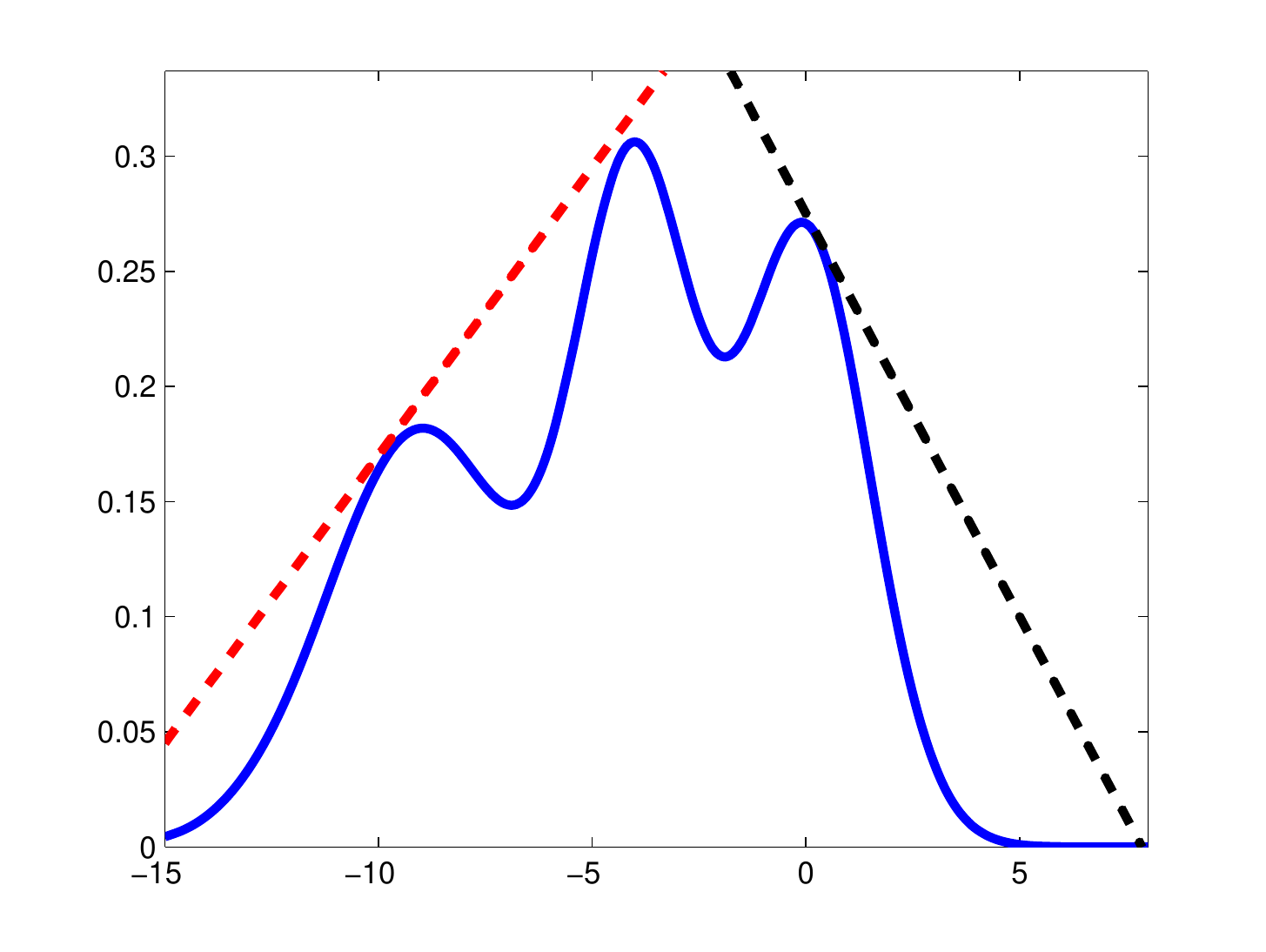}
\end{center}
\vskip -0.3in
\caption{Sampling as generalized BP.
}
\label{fig:sampling}
\vskip -0.2in
\end{figure}

\subsubsection{Bayesian Collaborative Deep Learning}
Besides the MAP estimates, a sampling-based algorithm for the Bayesian treatment of CDL is also proposed in \cite{CDL}. This algorithm turns out to be a Bayesian and generalized version of the well-known back-propagation (BP) learning algorithm. We list the key conditional densities as follows:

\textbf{For $\W+$}:
We denote the concatenation of $\W_{l,*n}$ and $\b_l^{(n)}$ as $\W_{l,*n}^+$. Similarly, the concatenation of $\X_{l,j*}$ and $1$ is denoted as $\X_{l,j*}^+$. The subscripts of $\I$ are ignored. Then
\begin{align*}
&p(\W_{l,*n}^+|\X_{l-1,j*},\X_{l,j*},\lambda_s)\\
\propto \  &\NM(\W_{l,*n}^+|0,\lambda_w^{-1}\I) \NM(\X_{l,*n}|\sigma(\X_{l-1}^+\W_{l,*n}^+),\lambda_s^{-1}\I).
\end{align*}

\textbf{For $\X_{l,j*}$ ($l\neq L/2$)}:
Similarly, we denote the concatenation of $\W_{l}$ and $\b_l$ as $\W_{l}^+$ and have
\begin{align*}
&p(\X_{l,j*}|\W_l^+,\W_{l+1}^+,\X_{l-1,j*},\X_{l+1,j*}\lambda_s)\\
\propto \ & \NM(\X_{l,j*}|\sigma(\X_{l-1,j*}^+\W_l^+),\lambda_s^{-1}\I)\cdot\\&\NM(\X_{l+1,j*}|\sigma(\X_{l,j*}^+\W_{l+1}^+),\lambda_s^{-1}\I).
\end{align*}
Note that for the last layer ($l=L$) the second Gaussian would be $\NM(\X_{c,j*}|\X_{l,j*},\lambda_s^{-1}\I)$ instead.

\textbf{For $\X_{l,j*}$ ($l= L/2$)}:
Similarly, we have
\begin{align*}
&p(\X_{l,j*}|\W_l^+,\W_{l+1}^+,\X_{l-1,j*},\X_{l+1,j*},\lambda_s,\lambda_v,\v_j)\\
\propto \ &\NM(\X_{l,j*}|\sigma(\X_{l-1,j*}^+\W_l^+),\lambda_s^{-1}\I)\cdot\\
&\NM(\X_{l+1,j*}|\sigma(\X_{l,j*}^+\W_{l+1}^+),\lambda_s^{-1}\I) \NM(\v_j|\X_{l,j*},\lambda_v^{-1}\I).
\end{align*}

\textbf{For $\v_j$}: The posterior $p(\v_j|\X_{L/2,j*},\R_{*j},\C_{*j},\lambda_v,\U)$
\begin{align*}
\propto\NM(\v_j|\X_{L/2,j*}^T,\lambda_v^{-1}\I)\prod\limits_i \NM(\R_{ij}|\u_i^T\v_j,\C_{ij}^{-1}).
\end{align*}

\textbf{For $\u_i$}: The posterior $p(\u_i|\R_{i*},\V,\lambda_u,\C_{i*})$
\begin{align*}
\propto\NM(\u_i|0,\lambda_u^{-1}\I)\prod\limits_j(\R_{ij}|\u_i^T\v_j|\C_{ij}^{-1}).
\end{align*}

Interestingly, if $\lambda_s$ goes to infinity and adaptive rejection Metropolis sampling (which involves using the gradients of the objective function to approximate the proposal distribution) is used, the sampling for $\W^+$ turns out to be a \emph{Bayesian generalized} version of BP. Specifically, as Figure~\ref{fig:sampling} shows, after getting the gradient of the loss function at one point (the red dashed line on the left), the next sample would be drawn in the region under that line, which is equivalent to a probabilistic version of BP. If a sample is above the curve of the loss function, a new tangent line (the black dashed line on the right) would be added to better approximate the distribution corresponding to the joint log-likelihood. After that, samples would be drawn from the region under both lines. During the sampling, besides searching for local optima using the gradients (MAP), the algorithm also takes the variance into consideration. That is why it is called \emph{Bayesian generalized back-propagation}.

%\begin{figure}[!tb]
%\begin{center}
%%\framebox[4.0in]{$\;$}
%%\includegraphics[height=5cm]{likeli1.eps}
%\includegraphics[height=4.0cm]{fig/sampling}
%\end{center}
%\vskip -0.3in
%\caption{Sampling as generalized BP.
%}
%\label{fig:sampling}
%\vskip -0.2in
%\end{figure}

\subsubsection{Marginalized Collaborative Deep Learning}
In SDAE, corrupted input goes through encoding and decoding to recover the clean input. Usually, different epochs of training use different corrupted versions as input. Hence generally, SDAE needs to go through enough epochs of training to see sufficient corrupted versions of the input. Marginalized SDAE (mSDAE) \cite{mSDAE} seeks to avoid this by marginalizing out the corrupted input and obtaining closed-form solutions directly. In this sense, mSDAE is more computationally efficient than SDAE.

As mentioned in \cite{li2015deep}, using mSDAE instead of the Bayesian SDAE could lead to more efficient learning algorithms. For example, in \cite{li2015deep}, the objective when using a one-layer mSDAE can be written as follows:
\begin{align}\label{eq:L_mSDAE}
\mathscr{L}=
&-\sum\limits_j\|\widetilde{\X}_{0,j*}\W_1-\overline{\X}_{c,j*}\|_2^2 -\sum\limits_{i,j}\frac{\C_{ij}}{2}(\R_{ij}-\u_i^T\v_j)^2 \nonumber\\
&-\frac{\lambda_u}{2}\sum\limits_i \|\u_i\|_2^2
-\frac{\lambda_v}{2}\sum\limits_j\|\v_j^T\P_1-\X_{0,j*}\W_1\|_2^2,
\end{align}
where $\widetilde{\X}_{0,j*}$ is the collection of $k$ different corrupted versions of ${\X}_{0,j*}$ (a $k$-by-$B$ matrix) and $\overline{\X}_{c,j*}$ is the $k$-time repeated version of ${\X}_{c,j*}$ (also a $k$-by-$B$ matrix). $\P_1$ is the transformation matrix for item latent factors.

The solution for $\W_1$ would be:
\begin{align*}
\W_1=E(\S_1)E(\Q_1)^{-1},
\end{align*}
where $\S_1=\overline{\X}_{c,j*}^T\widetilde{\X}_{0,j*}+\frac{\lambda_v}{2}\P_1^T\V\X_c$ and $\Q_1=\overline{\X}_{c,j*}^T\widetilde{\X}_{0,j*}+\frac{\lambda_v}{2}\X_c^T\X_c$. A solver for the expectation in the equation above is provided in \cite{mSDAE}. Note that this is a linear and one-layer case which can be generalized to the nonlinear and multi-layer case using the same techniques as in \cite{mSDAE,nmSDAE}.

As we can see, in marginalized CDL, the perception variables $\Om_p=\{\X_0,\X_c,\W_1\}$, the hinge variables $\Om_h=\{\V\}$, and the task variables $\Om_t=\{\P_1,\R,\U\}$.

\subsubsection{Collaborative Deep Ranking}
CDL assumes a collaborative filtering setting to model the ratings directly. However, the output of recommender systems is often a ranked list, which means it would be more natural to use ranking rather than ratings as the objective. With this motivation, collaborative deep ranking (CDR) is proposed \cite{yingcollaborative} to jointly perform representation learning and collaborative ranking. The corresponding generative process is the same as that of CDL except for Step \ref{enum:user} and \ref{enum:rating}, which should be replaced with:
\begin{compactitem}
\item For each user $i$,
\begin{compactenum}
\item Draw a latent user vector for each user $i$:
$$
\u_i \sim \NM(\0,\lambda_u^{-1}\I_K).
$$
\item For each pair-wise preference $(j,k)\in\mathcal{P}_i$, where $\mathcal{P}_i=\{(j,k):\R_{ij}-\R_{ik}>0\}$, draw the preference:
$
\De_{ijk} \sim \NM(\u_i^T\v_j-\u_i^T\v_k,\C_{ijk}^{-1}).\\
$
\end{compactenum}
\end{compactitem}

Following the generative process, the last term of Equation (\ref{eq:L}) becomes $-\sum\limits_{i,j,k}\frac{\C_{ijk}}{2}(\De_{ijk}-(\u_i^T\v_j-\u_i^T\v_k))^2$.
%\begin{align}\label{eq:L_ranking}
%\mathscr{L}=&-\frac{\lambda_u}{2}\sum\limits_i \|\u_i\|_2^2
%-\frac{\lambda_w}{2}\sum\limits_l(\|\W_l\|_F^2+\|\b_l\|_2^2)\nonumber\\&-\frac{\lambda_v}{2}\sum\limits_j\|\v_j-f_e(\X_{0,j*},\W^+)^T\|_2^2 \nonumber \\
%&-\frac{\lambda_n}{2}\sum\limits_j\|f_r(\X_{0,j*},\W^+)-\X_{c,j*}\|_2^2 \nonumber \\
%&-\sum\limits_{i,j,k}\frac{\C_{ijk}}{2}(\De_{ijk}-(\u_i^T\v_j-\u_i^T\v_k))^2.
%\end{align}
Similar algorithms can be used to learn the parameters in CDR. As reported in \cite{yingcollaborative}, using the ranking objective leads to significant improvement in the recommendation performance.

Following the definition in Section \ref{sec:general}, CDR's perception variables $\Om_p=\{\{\W_l\},\{\b_l\},\{\X_l\},\X_c\}$, the hinge variables $\Om_h=\{\V\}$, and the task variables $\Om_t=\{\U,\De\}$.

\subsubsection{Symmetric Collaborative Deep Learning}
Models like \cite{CDL,yingcollaborative} focus the deep learning component on modeling the item content. Besides the content information from the items, attributes of users sometimes contain much more important information. It is therefore desirable to extend CDL to model user attributes as well \cite{li2015deep}. We call this variant symmetric CDL. For example, using an extra mSDAE on the user attributes adds two extra terms in Equation (\ref{eq:L_mSDAE}), $-\frac{\lambda_u}{2}\sum\limits_i\|\u_i^T\P_2-\Y_{0,j*}\W_2\|_2^2$ and $-\sum\limits_i\|\widetilde{\Y}_{0,i*}\W_2-\overline{\Y}_{c,i*}\|_2^2$,
%\begin{align}\label{eq:L_user}
%\mathscr{L}=&-\frac{\lambda_v}{2}\sum\limits_j\|\v_j^T\P_1-\X_{0,j*}\W_1\|_2^2
%\nonumber \\
%&-\frac{\lambda_u}{2}\sum\limits_i\|\u_i^T\P_2-\Y_{0,j*}\W_2\|_2^2 \nonumber \\
%&-\sum\limits_j\|\widetilde{\X}_{0,j*}\W_1-\overline{\X}_{c,j*}\|_2^2
%-\sum\limits_i\|\widetilde{\Y}_{0,i*}\W_2-\overline{\Y}_{c,i*}\|_2^2 \nonumber \\
%&-\sum\limits_{i,j}\frac{\C_{ij}}{2}(\R_{ij}-\u_i^T\v_j)^2,
%\end{align}
where $\widetilde{\Y}_{0,j*}$ (a $k$-by-$D$ matrix for user attributes) is the collection of $k$ different corrupted versions of ${\Y}_{0,j*}$ and $\overline{\Y}_{c,i*}$ (also a $k$-by-$D$ matrix) is the $k$-time repeated version of ${\Y}_{c,i*}$ (the clean user attributes). $\P_2$ is the transformation matrix for user latent factors and $D$ is the number of user attributes. Similar to the marginalized CDL, the solution for $\W_2$ given other parameters is:
\begin{align*}
\W_2=E(\S_2)E(\Q_2)^{-1},
\end{align*}
where $\S_2=\overline{\Y}_{c,i*}^T\widetilde{\Y}_{0,i*}+\frac{\lambda_u}{2}\P_2^T\U\Y_c$ and $\Q_2=\overline{\Y}_{c,i*}^T\widetilde{\Y}_{0,i*}+\frac{\lambda_u}{2}\Y_c^T\Y_c$.

In symmetric CDL, the perception variables $\Om_p=\{\X_0,\X_c,\W_1,\Y_0,\Y_c,\W_2\}$, the hinge variables $\Om_h=\{\V,\U\}$, and the task variables $\Om_t=\{\P_1,\P_2,\R\}$.

\subsubsection{Discussion}\label{sec:discussion_recsys}
CDL is the first hierarchical Bayesian model to bridge the gap between state-of-the-art deep learning models and RS. By performing deep learning collaboratively, CDL and its variants can simultaneously extract an effective deep feature representation from the content and capture the similarity and implicit relationship between items (and users). This way, the perception component and the task-specific component are able to interact with each other to create synergy and further boost the recommendation accuracy. The learned representation may also be used for tasks other than recommendation. Unlike previous deep learning models which use a simple target such as classification \cite{DBLP:journals/acl/KalchbrennerGB14} and reconstruction \cite{DBLP:journals/jmlr/VincentLLBM10}, CDL-based models\footnote{During the review process of this paper, there are some newly published works based on BDL (e.g., some CDL-based works \cite{CRAE,DBLP:conf/kdd/ZhangYLXM16}).} use CF as a more complex target in a probabilistic framework.

As mentioned in Section \ref{sec:intro}, the synergy created by \emph{information exchange} between two components is crucial to the performance of BDL. In the CDL-based models above, the exchange is achieved by assuming Gaussian distributions that connect the hinge variables and the variables in the perception component (drawing the hinge variable $\v_j \sim \NM(\X_{\frac{L}{2},j*}^T,\lambda_v^{-1}\I_K)$ in the generative process of CDL, where $\X_{\frac{L}{2}}$ is a perception variable), which is simple but effective and efficient in computation. Among the five CDL-based models in Table \ref{table:summary}, three of them are HV models and the others are LV models, according to the definition in Section \ref{sec:general}. Since it has been verified that the HV CDL significantly outperforms its ZV counterpart \cite{CDL}, we can expect extra performance boosts from the LV counterparts of the three HV models.

Besides efficient \emph{information exchange}, the designs of the models also meet the i.i.d. requirement of the distribution concerning hinge variables discussed in Section \ref{sec:general} and are hence easily parallelizable. In some models to be introduced later, we will see alternative designs to enable efficient and i.i.d. information exchange between the two components of BDL.

\subsection{Bayesian Deep Learning for Topic Models}\label{sec:topic_models}
In this section, we review some examples of using BDL for topic models. These models combine the merits of PGM (which naturally incorporates the probabilistic relationships among variables) and NN (which learns deep representations efficiently), leading to significant performance boost.

\subsubsection{Relational Stacked Denoising Autoencoders as Topic Models}\label{sec:rsdae}
\textbf{Problem Statement and Notation}:
Assume we have a set of items~(articles or movies) $\X_c$, with $\X_{c,j*}^T \in \RB^B$ denoting the content (attributes) of item $j$. Besides, we use $\I_K$ to denote a $K$-dimensional identity matrix and $\S=[\s_1,\s_2,\cdots,\s_J]$ to denote the \emph{relational latent matrix} with $\s_j$ representing the relational properties of item $j$.
%\footnote{** Add notation for $\S$}

From the perspective of SDAE, the $J$-by-$B$
%\footnote{***Has the symbol $S$ been defined? ** Use B instead.}
matrix $\X_c$ represents the clean input to the SDAE and the noise-corrupted matrix of the same size is denoted by $\X_0$. Besides, we denote the output of layer $l$ of the SDAE, a $J$-by-$K_l$ matrix, by $\X_l$. Row $j$ of $\X_l$ is denoted by $\X_{l,j*}$, $\W_l$ and $\b_l$ are the weight matrix and bias vector of layer $l$, $\W_{l,*n}$ denotes column $n$ of $\W_l$, and $L$ is the number of layers. As a shorthand, we refer to the collection of weight matrices and biases in all layers as $\W^+$. Note that an $L/2$-layer SDAE corresponds to an $L$-layer network.

\textbf{Model Formulation}:
Here we will use the Bayesian SDAE introduced before as a building block for the relational stacked denoising autoencoder~(RSDAE) model.

\begin{figure} [tbp]
\begin{center}
  \begin{tabular}{ccc}
   \includegraphics*[height=30mm]{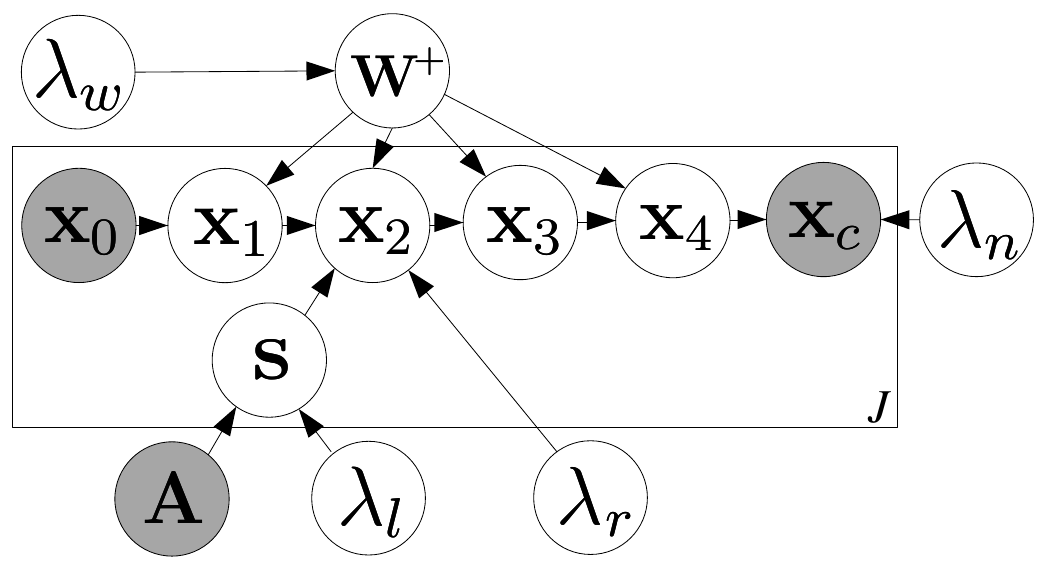}
  \end{tabular} \vskip -0.3cm
\caption{Graphical model of RSDAE for $L=4$. $\lambda_s$ is not shown here to prevent clutter.} \label{fig:rsdae}
\end{center}
\vskip -0.25in
\end{figure}

As mentioned in \cite{RSDAE}, RSDAE is formulated as a novel probabilistic model which can seamlessly integrate layered representation learning and the relational
%\footnote{*** Why not just stick to the term `relational'? ** All terms like network are replaced with relational if applicable.}
information available.  This way, the model can simultaneously learn the feature representation from the content information and the relation between items. The graphical model of RSDAE is shown in Figure~\ref{fig:rsdae} and the generative process is listed as follows:\\

\begin{compactenum}
\item Draw the relational latent matrix
%\footnote{*** Why do not suddenly change the terminology to social but not relational?  There is nothing `social' in our applications.  Also, it is better to define this term and notation earlier before you present the generative process that involves it. ** Changed to relational and add the notation earlier in the notation section.}
$\S$  from a \emph{matrix variate normal distribution}~\cite{gupta2000matrix}:
\begin{align}
\S\sim\NM_{K,J}(0,\I_K\otimes (\lambda_l \mathscr{L}_a)^{-1}).\label{eq:nm}
\end{align}
\item For layer $l$ of the SDAE where $l=1,2,\dots,\frac{L}{2}-1$,
\begin{compactenum}
\item For each column $n$ of the weight matrix $\W_l$, draw $\W_{l,*n} \sim \NM(0,\lambda_w^{-1} \I_{K_l})$.
\item Draw the bias vector $\b_l \sim \NM(0,\lambda_w^{-1} \I_{K_l})$.
\item For each row $j$ of $\X_l$, draw
$$
\X_{l,j*} \sim \NM(\sigma(\X_{l-1,j*}\W_l+\b_l),\lambda_s^{-1} \I_{K_l}).
$$
\end{compactenum}
\item For layer $\frac{L}{2}$ of the SDAE network, draw the representation vector for item $j$ from the product of two Gaussians~(PoG)~\cite{DBLP:journals/csl/GalesA06}:
\begin{align}
\X_{\frac{L}{2},j*}\sim\mbox{PoG}(\sigma(\X_{\frac{L}{2}-1,j*}\W_l+\b_l),\s_j^T,\lambda_s^{-1} \I_K,\lambda_r^{-1}\I_K). \label{eq:pog}
\end{align}
\item For layer $l$ of the SDAE network where $l=\frac{L}{2}+1,\frac{L}{2}+2,\dots,L$,
\begin{compactenum}
\item For each column $n$ of the weight matrix $\W_l$, draw $\W_{l,*n} \sim \NM(0,\lambda_w^{-1} \I_{K_l})$.
\item Draw the bias vector $\b_l \sim \NM(0,\lambda_w^{-1} \I_{K_l})$.
\item For each row $j$ of $\X_l$, draw
$$\X_{l,j*} \sim \NM(\sigma(\X_{l-1,j*}\W_l+\b_l),\lambda_s^{-1} \I_{K_l}).$$
\end{compactenum}
\item For each item $j$, draw a clean input $$\X_{c,j*} \sim \NM(\X_{L,j*},\lambda_n^{-1}\I_{B}).$$
\end{compactenum}
Here $K=K_{\frac{L}{2}}$ is the dimensionality of the learned representation vector for each item, $\S$ denotes the $K\times J$ relational latent matrix in which column $j$ is the \emph{relational latent vector} $\s_j$ for item $j$. Note that $\NM_{K,J}(0,\I_K\otimes (\lambda_l \mathscr{L}_a)^{-1})$ in~Equation (\ref{eq:nm})
%\footnote{*** Incorrect equation cited. ** Problem fixed.}
is a matrix variate normal distribution defined as in~\cite{gupta2000matrix}:
%\footnote{*** $p(\S)$ ** Modified.}
\begin{align}\label{eq:mvd}
p(\S) &= \NM_{K,J}(0,\I_K\otimes (\lambda_l \mathscr{L}_a)^{-1}) \nonumber  \\
 &=\frac{\exp\{\tr[-\frac{\lambda_l }{2}\S \mathscr{L}_a \S^T]\}}{(2\pi)^{JK/2}|\I_K|^{J/2}|\lambda_l \mathscr{L}_a|^{-K/2}},
\end{align}
where the operator $\otimes$ denotes the Kronecker product of two matrices \cite{gupta2000matrix}, $\tr(\cdot)$ denotes the trace of a matrix, and $\mathscr{L}_a$ is the Laplacian matrix incorporating the relational information.
%\footnote{*** Social network? ** changed to relational}
$\mathscr{L}_a=\D-\A$, where $\D$ is a diagonal matrix whose diagonal elements $\D_{ii}=\sum_j\A_{ij}$ and $\A$ is the adjacency matrix representing the relational information
%\footnote{** Changed.}
with binary entries indicating the links~(or relations) between items. $\A_{jj'}=1$ indicates that there is a link between item $j$ and item $j'$ and $\A_{jj'}=0$ otherwise. $\mbox{PoG}(\sigma(\X_{\frac{L}{2}-1,j*}\W_l+\b_l),\s_j^T,\lambda_s^{-1} \I_K,\lambda_r^{-1}\I_K)$ denotes the product of the Gaussian $\NM(\sigma(\X_{\frac{L}{2}-1,j*}\W_l+\b_l),\lambda_s^{-1} \I_K)$ and the Gaussian $\NM(\s_j^T,\lambda_r^{-1}\I_K)$, which is also a Gaussian \cite{DBLP:journals/csl/GalesA06}.
%The resulting Gaussian is $\NM(\mu_{sr},\lambda_{sr}^{-1}I_K)$ with
%\begin{align}
%\mu_{sr}&=\frac{\sigma(\X_{\frac{L}{2}-1,j*}\W_l+\b_l)\lambda_s+\s_j^T\lambda_r}{\lambda_v+\lambda_r}, \nonumber \\
%\lambda_{sr} &=\frac{\lambda_s\lambda_r}{\lambda_s+\lambda_r}. \nonumber
%\end{align}

According to the generative process above, maximizing the posterior probability is equivalent to maximizing the joint log-likelihood of $\{\X_l\}$, $\X_c$, $\S$, $\{\W_l\}$, and $\{\b_l\}$ given $\lambda_s$, $\lambda_w$, $\lambda_l$, $\lambda_r$, and $\lambda_n$:
\begin{align}
\mathscr{L}=&-\frac{\lambda_l}{2}\tr(\S\mathscr{L}_a \S^T)-\frac{\lambda_r}{2}\sum\limits_j\|(\s_j^T-\X_{\frac{L}{2},j*})\|_2^2\nonumber \\
&-\frac{\lambda_w}{2}\sum\limits_l(\|\W_l\|_F^2+\|\b_l\|_2^2) \nonumber \\
&-\frac{\lambda_n}{2}\sum\limits_j\|\X_{L,j*}-\X_{c,j*}\|_2^2 \nonumber \\
&-\frac{\lambda_s}{2}\sum\limits_l\sum\limits_j\|\sigma(\X_{l-1,j*}\W_l+\b_l)-\X_{l,j*}\|_2^2,\nonumber
\end{align}
%Similar to the generalized SDAE, taking $\lambda_s$ to infinity, the joint log-likelihood becomes:
%\begin{align}\label{eq:rsdae}
%\mathscr{L}=&-\frac{\lambda_l}{2}\tr(\S\mathscr{L}_a \S^T)-\frac{\lambda_r}{2}\sum\limits_j\|(\s_j^T-\X_{\frac{L}{2},j*})\|_2^2\nonumber \\
%&-\frac{\lambda_w}{2}\sum\limits_l(\|\W_l\|_F^2+\|\b_l\|_2^2) \nonumber \\
%&-\frac{\lambda_n}{2}\sum\limits_j\|\X_{L,j*}-\X_{c,j*}\|_2^2,
%\end{align}
where $\X_{l,j*}=\sigma(\X_{l-1,j*}\W_l+\b_l)$. Note that the first term $-\frac{\lambda_l}{2}\tr(\S\mathscr{L}_a \S^T)$ corresponds to $\log p(\S)$
in the matrix variate distribution in Equation (\ref{eq:mvd}).
By simple manipulation, we have
%\begin{align}
%\tr(\S \mathscr{L}_a \S^T) &=\frac{1}{2}\sum\limits_{j=1}^J\sum\limits_{j'=1}^J \A_{jj'}||\S_{*j}-\S_{*j'}||^2 \\ \nonumber
%&=\frac{1}{2}\sum\limits_{j=1}^J\sum\limits_{j'=1}^J[\A_{jj'}\sum\limits_{k=1}^K(\S_{kj}-\S_{kj'})^2]\\ \nonumber
%&=\frac{1}{2}\sum\limits_{k=1}^{K}[\sum\limits_{j=1}^J\sum\limits_{j'=1}^J \A_{jj'}(\S_{kj}-\S_{kj'})^2]\\ \nonumber
%&=\sum\limits_{k=1}^K \S_{k*}^T\mathscr{L}_a\S_{k*}, \nonumber
%\end{align}
$
\tr(\S \mathscr{L}_a \S^T)
=\sum\limits_{k=1}^K \S_{k*}^T\mathscr{L}_a\S_{k*}
$,
where $\S_{k*}$ denotes the $k$-th row of $\S$. As we can see, maximizing $-\frac{\lambda_l}{2}\tr(\S^T\mathscr{L}_a\S)$ is equivalent to making $\s_j$ closer to $\s_{j'}$ if item $j$ and item $j'$ are linked (namely $\A_{jj'}=1$).

In RSDAE, the perception variables $\Om_p=\{\{\X_l\},\X_c,\{\W_l\},\{\b_l\}\}$, the hinge variables $\Om_h=\{\S\}$, and the task variables $\Om_t=\{\A\}$.

\textbf{Learning Relational Representation and Topics}:
\cite{RSDAE} provides an EM-style algorithm for MAP estimation. Here we review some of the key steps as follows.

In terms of the relational latent matrix $\S$,
%\footnote{*** $S$ is not in boldface from here onwards. ** Modified.}
we first fix all rows of $\S$ except the $k$-th one $\S_{k*}$ and then update $\S_{k*}$. Specifically, we take the gradient of $\mathscr{L}$ with respect to $\S_{k*}$, set it to 0, and get the following linear system:
\begin{align}
(\lambda_l\mathscr{L}_a+\lambda_r\I_J)\S_{k*}=\lambda_r\X_{\frac{L}{2},*k}^T.
\end{align}
\begin{table}[!tb]
\begin{scriptsize}
\centering
\vskip -0.1cm
\caption{Recall@300 on the dataset \emph{movielens-plot} ($\%$)}\label{table:rsdae_recall}
\vskip -0.2cm
\begin{tabular}{|c|c|c|c|} \hline
CTR \cite{CTR} & CTR-SR \cite{CTRSR} & SDAE \cite{DBLP:journals/jmlr/VincentLLBM10} & RSDAE \cite{RSDAE} \\ \hline
20.43 &23.07	&23.45&	\textbf{24.86} \\ \hline
\end{tabular}
\vskip -0.5cm
\end{scriptsize}
\end{table}
A naive approach is to solve the linear system by setting $\S_{k*}=\lambda_r(\lambda_l\mathscr{L}_a+\lambda_r\I_J)^{-1}\X_{\frac{L}{2},*k}^T$. Unfortunately, the complexity is $O(J^3)$ for one single update. Similar to \cite{DBLP:conf/ijcai/LiY09}, the steepest descent method \cite{techreport/Shewchuk94} is used to iteratively update $\S_{k*}$:
\begin{align}
\S_{k*}(t+1)&\leftarrow \S_{k*}(t)+\delta(t)r(t)\nonumber\\
r(t)&\leftarrow \lambda_r\X_{\frac{L}{2},*k}^T-(\lambda_l\mathscr{L}_a+\lambda_r\I_J)\S_{k*}(t)\nonumber\\
\delta(t)&\leftarrow \frac{r(t)^Tr(t)}{r(t)^T(\lambda_l\mathscr{L}_a+\lambda_r\I_J)r(t)}.\nonumber
\end{align}
As discussed in \cite{DBLP:conf/ijcai/LiY09}, the use of steepest descent method dramatically reduces the computation cost in each iteration from $O(J^3)$ to $O(J)$.

Given $\S$, we can learn $\W_l$ and $\b_l$ for each layer using the back-propagation algorithm. By alternating the update of $\S$, $\W_l$, and $\b_l$, a local optimum for $\mathscr{L}$ can be found. Also, techniques such as including a momentum term may help to avoid being trapped in a local optimum. The computational complexity for each iteration is $O(JBK_1+KJ)$. Similar to CDL, no significant overhead is introduced.

Table \ref{table:rsdae_recall} shows the recall for different methods in the dataset \emph{movielens-plot} when the learned representation is used for tag recommendation (with $300$ recommended tags for each item). As we can see, RSDAE significantly outperforms SDAE, which means that the relational information from the task-specific component is crucial to the performance boost. Please refer to \cite{RSDAE} for more details.

\subsubsection{Deep Poisson Factor Analysis with Sigmoid Belief Networks}\label{sec:dpfa_sbn}
The Poisson distribution with support over nonnegative integers is known as a natural choice to model counts. It is, therefore, desirable to use it as a building block for topic models \cite{LDA}. With this motivation, \cite{PFA} proposed a model, dubbed Poisson factor analysis (PFA), for latent nonnegative matrix factorization via Poisson distributions.

\textbf{Poisson Factor Analysis}:
PFA assumes a discrete $N$-by-$P$ matrix $\X$ containing word counts of $N$ documents with a vocabulary size of $P$ \cite{PFA,DPFA}. In a nutshell, PFA can be described using the following equation:
\begin{align}
\X\sim \mbox{Pois}((\Tha\circ\H)\Ph), \label{eq:pfa}
\end{align}
where $\Ph$ (of size $K$-by-$P$ where $K$ is the number of topics) denotes the factor loading matrix in factor analysis with the $k$-th row $\ph_k$ encoding the importance of each word in topic $k$. The $N$-by-$K$ matrix $\Tha$ is the factor score matrix with the $n$-th row $\tha_n$ containing topic proportions for document $n$. The $N$-by-$K$ matrix $\H$ is a latent binary matrix with the $n$-th row $\h_n$ defining a set of topics associated with document $n$. %Note that the superscript $(1)$ indexes a layer, which is irrelevant for one-layer PFA.

Different priors correspond to different models. For example, Dirichlet priors on $\ph_k$ and $\tha_n$ with an all-one matrix $\H$ would recover LDA \cite{LDA} while a beta-Bernoulli prior on $\h_n$ leads to the negative binomial focused topic model (NB-FTM) model in \cite{DBLP:journals/pami/ZhouC15}. In \cite{DPFA}, a deep-structured prior based on sigmoid belief networks (SBN) \cite{neal1992} (an MLP variant with binary hidden units) is imposed on $\h_n$ to form a deep PFA model for topic modeling.

\textbf{Deep Poisson Factor Analysis}:
In the deep PFA model \cite{DPFA}, the generative process can be summarized as follows:
\begin{align}
\ph_k&\sim\mbox{Dir}(a_{\phi},\dots,a_{\phi}),\theta_{nk}\sim\mbox{Gamma}(r_k,\frac{p_n}{1-p_n})\nonumber\\
r_k&\sim\mbox{Gamma}(\gamma_0,\frac{1}{c_0}),\gamma_0\sim\mbox{Gamma}(e_0,\frac{1}{f_0})\nonumber\\
h_{nk_L}^{(L)}&\sim\mbox{Ber}(\sigma(b_{k_L}^{(L)}))\label{eq:sbn_top}\\
h_{nk_l}^{(l)}&\sim\mbox{Ber}(\sigma(\h_n^{(l+1)}\w_{k_l}^{(l)}+b_{k_l}^{(l)}))\label{eq:sbn_others}\\
x_{npk}&\sim\mbox{Pois}(\phi_{kp}\theta_{nk}h_{nk}^{(1)}),x_{np}=\sum\limits_{k=1}^Kx_{npk}, \label{eq:sbn_x}
\end{align}
%\begin{align}
%\ph_k&\sim\mbox{Dir}(a_{\phi},\dots,a_{\phi}),\theta_{nk}\sim\mbox{Gamma}(r_k,\frac{p_n}{1-p_n})\nonumber\\
%r_k&\sim\mbox{Gamma}(\gamma_0,\frac{1}{c_0}),\gamma_0\sim\mbox{Gamma}(e_0,\frac{1}{f_0})\nonumber\\
%h_{nk_L}^{(L)}&\sim\mbox{Ber}(\sigma(b_{k_L}^{(L)}))\label{eq:sbn_top}\\
%h_{nk_l}^{(l)}&\sim\mbox{Ber}(\sigma(\h_n^{(l+1)}\w_{k_l}^{(l)}+b_{k_l}^{(l)}))\label{eq:sbn_others}\\
%x_{npk}&\sim\mbox{Pois}(\phi_{kp}\theta_{nk}h_{nk}^{(1)}) \label{eq:sbn_x} \\
%x_{np}&=\sum\limits_{k=1}^Kx_{npk},\nonumber
%\end{align}
where $L$ is the number of layers in SBN, which corresponds to Equation (\ref{eq:sbn_top}) and (\ref{eq:sbn_others}). $x_{np}$ is an entry in the matrix $\X$, $\h_n^{(l)}$ is the $n$-th row of $\H_l$, and $x_{npk}$ is the count of word $p$ that comes from topic $k$ in document~$n$.

In this model, the perception variables $\Om_p=\{\{\H_l\},\{\W_l\},\{\b_l\}\}$, the hinge variables $\Om_h=\{\X\}$, and the task variables $\Om_t=\{\{\ph_k\},\{r_k\},\Tha,\gamma_0\}$. $\W_l$ is the weight matrix containing columns of $\w_{k_l}^{(l)}$ and $\b_l$ is the bias vector containing entries of $b_{k_l}^{(l)}$ in Equation (\ref{eq:sbn_others}).

\textbf{Learning Using Bayesian Conditional Density Filtering}:
Efficient learning algorithms are needed for Bayesian treatments of deep PFA. \cite{DPFA} proposed to use an online version of MCMC called Bayesian conditional density filtering (BCDF) to learn both the global parameters $\Ps_g=(\{\ph_k\},\{r_k\},\gamma_0,\{\W_l\},\{\b_l\})$ and the local variables $\Ps_l=(\Tha,\{\H_l\})$. The key conditional densities used for the Gibbs updates are as follows:
\begin{align*}
x_{npk}|-&\sim\mbox{Multi}(x_{np};\zeta_{np1},\dots,\zeta_{npK})\\
\ph_k|-&\sim\mbox{Dir}(a_{\phi}+x_{\cdot 1 k},\dots,a_{\phi}+x_{\cdot P k})\\
\theta_{nk}|-&\sim\mbox{Gamma}(r_kh_{nk}^{(1)}+x_{n \cdot k},p_n)\\
h_{nk}^{(1)}|-&\sim \delta(x_{n \cdot k}=0)\mbox{Ber}(\frac{\widetilde{\pi}_{nk}}{\widetilde{\pi}_{nk}+(1-\pi_{nk})})+\delta(x_{n \cdot k}>0),
\end{align*}
where $\widetilde{\pi}_{nk}=\pi_{nk}(1-p_n)^{r_k}$, $\pi_{nk}=\sigma(\h_n^{(2)}\w_k^{(1)}+b_k^{(1)})$, $x_{n \cdot k}=\sum\limits_{p=1}^P x_{npk}$, $x_{ \cdot p k}=\sum\limits_{n=1}^N x_{npk}$, and $\zeta_{npk}\propto\phi_{kp}\theta_{nk}$. For the learning of $h_{nk}^{(l)}$ where $l>1$, the same techniques as in \cite{DBLP:conf/aistats/GanHCC15} can be used.

\textbf{Learning Using Stochastic Gradient Thermostats}:
An alternative way of learning deep PFA is through the use of \emph{stochastic gradient N\'ose-Hoover thermostats} (SGNHT), which is more accurate and scalable.
%SGNHT is a generalization of the \emph{stochastic gradient Langevin dynamics} (SGLD) \cite{SGLD} and the \emph{stochastic gradient Hamiltonian Monte Carlo} (SGHMC) \cite{SGHMC}. Compared with the previous two, it introduces momentum variables into the system, helping the system to jump out of local optima.
Specifically, the following stochastic differential equations (SDE) can be used:
\begin{align*}
d\Ps_g&=\v dt,d\v=\widetilde{f}(\Ps_g)dt-\xi\v dt+\sqrt{D}d\mathcal{W}\\
d\xi&=(\frac{1}{M}\v^T\v-1)dt,
\end{align*}
%\begin{align*}
%d\Ps_g&=\v dt\\
%d\v&=\widetilde{f}(\Ps_g)dt-\xi\v dt+\sqrt{D}d\mathcal{W}\\
%d\xi&=(\frac{1}{M}\v^T\v-1)dt,
%\end{align*}
where $\widetilde{f}(\Ps_g)=-\nabla_{\Ps_g}\widetilde{U}(\Ps_g)$ and $\widetilde{U}(\Ps_g)$ is the negative log-posterior of the model. $t$ indexes time and $\mathcal{W}$ denotes the standard Wiener process. $\xi$ is the thermostats variable to make sure the system has a constant temperature. $D$ is the injected variance which is a constant. %To speed up convergence, the SDE is generalized to:
%\begin{align*}
%d\Ps_g&=\v dt\\
%d\v&=\widetilde{f}(\Ps_g)dt-\Xii\v dt+\sqrt{D}d\mathcal{W}\\
%d\Xii&=(\q-\I)dt,
%\end{align*}
%where $\I$ is the identity matrix, $\Xii=\mbox{diag}(\xi_1,\dots,\xi_M)$, $\q=\mbox{diag}(v_1^2,\dots,v_M^2)$, and $M$ is the dimensionality of the parameters.

\subsubsection{Deep Poisson Factor Analysis with Restricted Boltzmann Machine}
Similar to the deep PFA above, the restricted Boltzmann machine (RBM) \cite{RBM} can be used in place of SBN \cite{DPFA}. If RBM is used, Equation (\ref{eq:sbn_top}) and (\ref{eq:sbn_others}) would be defined using the energy \cite{RBM}:
\begin{align*}
E(\h_n^{(l)},\h_n^{(l+1)})=&-\h_n^{(l)}\b_l^T-\h_n^{(l)}\W^{(l)}{\h_n^{(l+1)}}^T\\
&-\h_n^{(l+1)}\b_{l+1}^T.
\end{align*}

For the learning, similar algorithms as the deep PFA with SBN can be used. Specifically, the sampling process would alternate between $\{\{\ph_k\},\{\gamma_k\},\gamma_0\}$ and $\{\{\W_l\},\{\b_l\}\}$. For $\{\{\ph_k\},\{\gamma_k\},\gamma_0\}$, similar conditional density as the SBN-based DPFA is used. For $\{\{\W_l\},\{\b_l\}\}$, they use the \emph{contrastive divergence} algorithm.

\subsubsection{Discussion}\label{sec:discussion_topic_models}
In BDL-based topic models, the perception component is responsible for inferring the topic hierarchy from documents while the task-specific component is in charge of modeling the word generation, topic generation, word-topic relation, or inter-document relation. The synergy between these two components comes from the bidirectional interaction between them. On the one hand, knowledge of the topic hierarchy would facilitate accurate modeling of words and topics, providing valuable information for learning inter-document relations. On the other hand, accurately modeling the words, topics, and inter-document relations could help with the discovery of topic hierarchy and learning of compact document representations.

As we can see, the \emph{information exchange} mechanism in some BDL-based topic models is different from that in Section \ref{sec:recsys}. For example, in the SBN-based DPFA model, the exchange is natural since the bottom layer of SBN, $\H_1$, and the relationship between $\H_1$ and $\Om_h=\{\X\}$ are both inherently probabilistic, as shown in Equation (\ref{eq:sbn_others}) and (\ref{eq:sbn_x}), which means additional assumptions about the distribution are not necessary. The SBN-based DPFA model is equivalent to assuming that $\H$ in PFA (see Equation (\ref{eq:pfa})) is generated from a Dirac delta distribution (a Gaussian distribution with zero variance) centered at the bottom layer of the SBN, $\H_1$. Hence both DPFA models in Table \ref{table:summary} are ZV models, according to the definition in Section \ref{sec:general}. It is worth noting that RSDAE is an HV model (see Equation (\ref{eq:pog}), where $\S$ is the hinge variable and the others are perception variables), and naively modifying this model to be its ZV counterpart would violate the i.i.d. requirement in Section~\ref{sec:general}.

\subsection{Other Applications}\label{sec:other}
As mentioned in Section \ref{sec:intro}, BDL can also be applied to applications beyond data engineering and data mining (e.g., the control of nonlinear dynamical systems from raw images or medical diagnosis with medical images).

Consider controlling a complex dynamical system according to the live video stream received from a camera. One way of solving this control problem is by iteration between two tasks, perception from raw images and control based on dynamic models. The perception task can be taken care of using multiple layers of simple nonlinear transformation (deep learning) while the control task usually needs more sophisticated models such as hidden Markov models and Kalman filters \cite{harrison1999bayesian,DBLP:conf/uai/MatsubaraGK14}. To enable an effective iterative process between the perception task and the control task, two-way information exchange between them is often necessary. The perception component would be the basis on which the control component estimates its states and on the other hand, the control component with a built-in dynamic model would be able to predict the future trajectory (images) by reversing the perception process. For example, \cite{watter2015embed} proposed a BDL-based model that performs control based on the received raw images (videos). Their key generative process is as follows:
\begin{align}
\z_t&\sim& Q_{\phi}(Z|X)&=\NM(\muu_t,\Si_t) \nonumber\\
\widetilde{\z}_{t+1}&\sim& \widetilde{Q}_{\psi}(\widetilde{Z}|Z,\u)&=\NM(\A_t\muu_t+\B_t\u_t+\oo_t,\C_t) \nonumber\\
\widetilde{\x}_t,\widetilde{\x}_{t+1}&\sim& P_{\theta}(X|Z)&=Bernoulli(\p_t), \nonumber
\end{align}
where $Q_{\phi}(Z|X)$ is the encoding model which encodes the raw images $X$ into latent states $Z$. $\widetilde{Q}_{\psi}(\widetilde{Z}|Z,\u)$ is the transition model which predicts the next latent state $\widetilde{Z}$ given the current latent state $Z$ and the applied control $\u$. $P_{\theta}(X|Z)$ is the reconstruction (decoding) model which reconstructs the raw images $X$ from latent states $Z$. The parameters $\muu_t$, $\Si_t$, $\A_t$, $\B_t$, $\oo_t$, $\C_t$, and $\p_t$ are then further parameterized by neural networks.

It is worth noting that in terms of \emph{information exchange} between the two components, this BDL-based control model uses a different mechanism from the ones in Section \ref{sec:recsys} and Section \ref{sec:topic_models}: it uses neural networks to \emph{separately} parameterize the mean and covariance of hinge variables (e.g., $\muu_t$ and $\Si_t$ in the encoding model), which is more flexible (with more free parameters) than models such as CDL and CDR in Section \ref{sec:recsys}, where Gaussian distributions with fixed variance are also used. Note that this BDL-based control model is an LV model, and since the covariance is assumed to be diagonal \cite{watter2015embed}, the model still meets the i.i.d. requirement in Section \ref{sec:general}.

%\section{Future Research}\label{sec:future}
\section{Conclusions and Future Research}\label{sec:summary}
In this paper, we identified a current trend of merging probabilistic graphical models and neural networks (deep learning), proposed a BDL framework, and reviewed relevant recent work on BDL, which strives to combine the merits of PGM and NN by organically integrating them in a single principled probabilistic framework. To learn parameters in BDL, several algorithms have been proposed, ranging from block coordinate descent, Bayesian conditional density filtering, and stochastic gradient thermostats to stochastic gradient variational Bayes.

BDL has gained its popularity both from the success of PGM and from recent promising advances in deep learning. Since many real-world tasks involve both perception and inference, BDL is a natural choice for harnessing the perception ability from NN and the (causal and logical) inference ability from PGM. Although current applications of BDL focus on recommender systems, topic models, and stochastic optimal control, in the future, we can expect an increasing number of other applications such as link prediction, community detection, active learning, Bayesian reinforcement learning, and many other complex tasks that need interaction between perception and causal inference. In these complex tasks, BDL with interconnected perception components (to handle perception) and task-specific components (to handle inference/reasoning) possesses great performance-boosting potential. Besides, with the advances of efficient \emph{Bayesian neural networks} (BNN), BDL with BNN as an important component is expected to be more and more scalable.

\bibliography{pqe,hao,bnn,bdl,nn}
\bibliographystyle{plain}


% deleted by hog
%\end{IEEEbiography}
%% insert where needed to balance the two columns on the last page with
%%\newpage
\begin{IEEEbiography}{Hao Wang}
received his B.Sc. degree in Computer Science from Shanghai Jiao Tong University, China. He is currently a Ph.D student at the Department of Computer Science and Engineering of Hong Kong University of Science and Technology. His research interests are in statistical machine learning and data mining. He received the Hong Kong PhD Fellowship in 2013, the Microsoft Research Asia Fellowship in 2015, and the Baidu Research Fellowship in 2015 for his achievements on statistical machine learning and Bayesian deep learning.
\end{IEEEbiography}

\begin{IEEEbiography}{Dit-Yan Yeung}
received his BEng degree in electrical engineering and MPhil degree in computer science from the University of Hong Kong, and PhD degree in computer science from the University of Southern California.  He started his academic career as an assistant professor at the Illinois Institute of Technology in Chicago.  He then joined the Hong Kong University of Science and Technology where he is now a full professor in the Department of Computer science and Engineering, with joint appointment in the Department of Electronic and Computer Engineering.  His research interests are in computational and statistical approaches to machine learning and artificial intelligence.
\end{IEEEbiography}


%\vfill


\end{document}